\documentclass[10pt,conference]{IEEEtran}
\IEEEoverridecommandlockouts
\usepackage[font=small,skip=3pt]{caption}
\usepackage{enumitem}
\usepackage{subcaption}
\usepackage{booktabs} 
\usepackage{bbm}
\usepackage{cite}
\usepackage{adjustbox} 
\usepackage{graphicx}  
\usepackage{array} 
\usepackage{makecell} 
\usepackage{pifont}   
\usepackage{array}
\usepackage{url}
\usepackage{caption}
\newcolumntype{R}[1]{%
  >{\adjustbox{angle=90,lap=\width-\height}}%
  m{#1}%
}
\usepackage{setspace}
\usepackage{multirow}
\usepackage{geometry}
\newcommand{\cmark}{\ding{51}}%
\newcommand{\xmark}{\ding{55}}%
\usepackage{amsthm} 
\usepackage{amsmath} 
\usepackage{amssymb}
\newtheorem{theorem}{Theorem}

\usepackage[ruled,vlined,linesnumbered]{algorithm2e}
\usepackage[table,xcdraw,dvipsnames]{xcolor} % For coloring
\SetKwInput{KwData}{Input}
\SetKwInput{KwResult}{Output}
\usepackage{mathtools}

\DeclareRobustCommand{\rchi}{{\mathpalette\irchi\relax}}
\newcommand{\irchi}[2]{\raisebox{\depth}{$#1\chi$}}

\usepackage{fancyhdr}
\begin{document}

% \title{Charging and Routing Co-Optimization for Last Mile Deliveries with Electric Vehicles}
\title{CARGO: A Co-Optimization Framework for EV Charging and Routing in Goods Delivery Logistics}
% \title{Energy and Time-Constrained Last-Mile  Deliveries Using EV: A Cost-Effective Approach}
% \title{Efficient Last-mile Deliveries using EV}

\author{
\IEEEauthorblockN{
Arindam Khanda\IEEEauthorrefmark{1}\IEEEauthorrefmark{3}, 
Anurag Satpathy\IEEEauthorrefmark{1}\IEEEauthorrefmark{3}, 
Amit Jha\IEEEauthorrefmark{2}, 
Sajal K. Das\IEEEauthorrefmark{1}
}
\IEEEauthorblockA{\IEEEauthorrefmark{1}Department of Computer Science, Missouri University of Science and Technology, Rolla, MO, USA \\
Email: \{akkcm, anurag.satpathy, sdas\}@mst.edu}
\IEEEauthorblockA{\IEEEauthorrefmark{2}CGI, USA \\
Email: amit.n.jha@cgi.com}
\IEEEauthorblockA{\IEEEauthorrefmark{3}These authors contributed equally to this work.}
}

\maketitle

\begin{abstract}
With growing interest in sustainable logistics, electric vehicle (EV)-based deliveries offer a promising alternative for urban distribution. However, EVs face challenges due to their limited battery capacity, requiring careful planning for recharging. This depends on factors such as the charging point (CP) availability, cost, proximity, and vehicles' state of charge (SoC). We propose CARGO, a framework addressing the EV-based delivery route planning problem (EDRP), which jointly optimizes route planning and charging for deliveries within time windows. After proving the problem's NP-hardness, we propose a mixed integer linear programming (MILP)-based exact solution and a computationally efficient heuristic method. Using real-world datasets, we evaluate our methods by comparing the heuristic to the MILP solution, and benchmarking it against baseline strategies, Earliest Deadline First (EDF) and Nearest Delivery First (NDF). The results show up to 39\% and 22\% reductions in the charging cost over EDF and NDF, respectively, while completing comparable deliveries.

\end{abstract}

\thispagestyle{empty}
% \vspace{-0.05in}
\section{Introduction}\label{sec:introduction}
% \vspace{-0.05in}
Delivery systems form the backbone of modern logistics, facilitating the movement of goods across regional, inter-city, and urban networks \cite{galic2013case}. These systems face increasing pressure to remain cost-efficient, responsive, and scalable amid growing demand for fast, flexible services. 
Interestingly, transportation-related activities account for over $60\%$ of total logistics expenditures~\cite{cscmp2023}, highlighting the critical role of delivery operations in overall supply chain performance. Large-scale providers such as \texttt{UPS}, \texttt{Maersk}, and \texttt{XPO Logistics} manage scheduled, multi-modal freight across warehouses, retailers, and consumers, focusing on reliability and capacity utilization. In contrast, urban delivery platforms support rapid, customer-facing fulfillment. For instance, \texttt{Amazon Prime Now} delivers groceries within a $2$-\textit{hour} window in select cities, while services like \texttt{DoorDash}, \texttt{Uber Eats}, and \texttt{Instacart} complete deliveries in \textit{30–60}~\textit{minutes}. Courier companies such as \texttt{FedEx} and \texttt{DHL Express} offer customizable, time-sensitive shipping across consumer and enterprise markets.
% Last-mile delivery is the final phase of the delivery process, where goods move from a distribution hub or warehouse to the customer's location \cite{zhang2023stable}. It is the most complex and costly operation in the supply chain, comprising 28\% of transportation costs and 41\% of overall supply chain expenses \cite{jacobs2019last}. The rapid growth of e-commerce has introduced operational challenges, particularly the rising demand for fast, responsive deliveries \cite{jacobs2019last}. For example, \texttt{Amazon Prime Now} delivers groceries and essentials within a 2-\textit{hour} window in select cities. Similarly, platforms like \texttt{DoorDash}, \texttt{Uber Eats}, and \texttt{Instacart} offer food and grocery deliveries in $30$ to $60$ \textit{minutes}. Meanwhile, \texttt{FedEx} and \texttt{DHL Express} offer customizable, time-sensitive shipping for fast, seamless service.
\begin{figure}
    \centering
    % \vspace{-0.1in}
    \includegraphics[width=0.94\columnwidth]{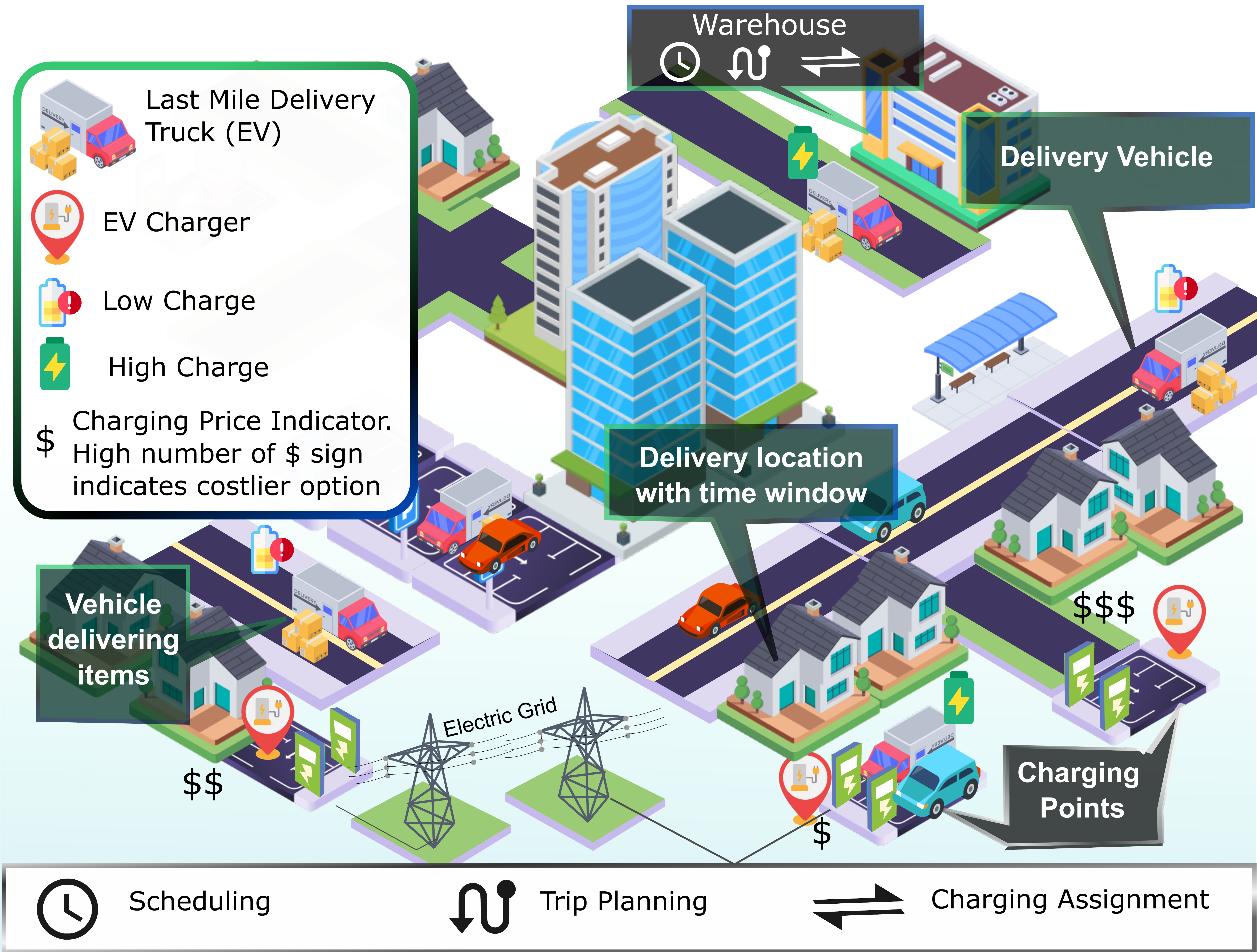}
    \caption{Schematic view of a goods delivery system in an urban setting.}
    % \vspace{-0.1in}
    \label{fig:IntroModel}
\end{figure}
% \vspace{-0.02in}

The global shift toward EVs is accelerating, gradually replacing internal combustion engine (ICE) vehicles and reducing reliance on fossil fuels. Governments worldwide support this transition through regulatory measures and investment \cite{9843207, 9524937}. For example, \texttt{Canada} and the \texttt{United Kingdom} have announced plans to phase out ICE vehicles by \textit{2040}, \texttt{China} suspended new ICE investments in \textit{2019}, and \texttt{California} has set a target of $5$~\textit{million} EVs by \textit{2030}. Global EV adoption is projected to exceed \textit{250}~\textit{million} units by the same year~\cite{shurrab2022stable}. This transformation is reshaping delivery system design and operations, which have traditionally been optimized for ICE-powered fleets or, in some cases, unmanned aerial vehicles (UAVs)~\cite{moshref2020truck, 9637806}. The integration of EVs into delivery fleets is gaining momentum, with the market for EV-based delivery solutions projected to grow at a compound annual growth rate (CAGR) of $15.6\%$ between \textit{2023} and \textit{2028}~\cite{GlobalMarketEstimates2023}. Consumer expectations are evolving in parallel: a \textit{2023} survey showed that $31\%$ of online shoppers support the immediate deployment of EV delivery fleets—up from $20\%$ in \textit{2022}—while $61\%$ expect widespread adoption within $5$ years~\cite{Statista2023}. Industry leaders are responding; \texttt{Amazon}, for instance, has committed to deploying $100{,}000$ EV delivery vehicles by $2030$, with over $5{,}000$ \texttt{Rivian} EVs already operational as of July \textit{2023}~\cite{Wikipedia2025}. 
% Beyond environmental benefits, EV-based delivery systems could reduce vehicle miles traveled by $34\%$–$56\%$ and energy consumption by $29\%$–$54\%$ compared to conventional delivery fleets~\cite{AFDC2025}. 
However, this transition presents several open research challenges. Limited charging infrastructure \cite{khanda2025smevca}, extended charging times \cite{EPRINC2024}, and shorter driving ranges compared to ICE vehicles \cite{hardman2018battery}, thereby requiring careful routing and energy management coordination to sustain operational efficiency in large-scale delivery systems.

To address the challenges outlined above, this paper presents \textsc{CARGO}, a delivery route planning framework that solves the Electric Vehicle-based Delivery Route Planning (EDRP) problem by jointly optimizing: (1) delivery routing for timely, energy- and cost-efficient operations, and (2) strategic charging decisions to manage battery constraints. While prior work has typically addressed either routing~\cite{suh2012leveraging, han2017appointment, moshref2020truck, danchuk2021simulation, danchuk2023optimization, alvarez2024optimizing} or charging~\cite{algafri2024smart, sassi2017electric, tao2023distributed, elghitani2020efficient, bus-coop} in isolation, CARGO integrates both, enabling coordinated decisions that minimize operational costs while meeting strict delivery time windows. The framework provides two solution approaches: an exact method based on MILP and a heuristic offering a tractable alternative with improved computational efficiency. While co-optimization using MILP significantly reduces costs compared to sequential methods, it increases complexity due to the enlarged solution space—the heuristic addresses this by providing a practical, scalable alternative. The \textbf{novel contributions} are summarized as follows:

\begin{enumerate}[label=(\arabic*)] 
\item We propose \textsc{CARGO} to solve the EDRP problem by integrating charging decisions into delivery routing, considering station availability, costs, proximity, SoC, and associated loading, charging, and waiting times.
\item The problem is formulated as a Mixed-Integer Linear Programming (MILP) model that jointly optimizes routing and charging decisions to minimize delivery and charging costs, subject to constraints on charging schedules, delivery time windows, and trip limits.
\item We propose two solution approaches: an exact solver-based method that guarantees optimality but is computationally intensive, and a heuristic that offers faster execution at the cost of solution optimality.
\item For experimentation, we use the real-world dataset extracted from \texttt{JD Logistics}’ distribution system \cite{zheng2024hybrid}.
Performance validation is conducted in two phases: first, comparing the optimal solution (Optimal), implemented with \texttt{Gurobi} 
% on a high-performance computing (HPC) cluster,
against the heuristic; second, benchmarking the heuristic against the Earliest Deadline First (EDF) and Nearest Delivery First (NDF) baselines.
The results demonstrated a $\gtrapprox39\%$ and $\gtrapprox22\%$ reduction in charging costs compared to EDF and NDF, respectively, while maintaining a comparable number of successful deliveries. Additionally, the implementation of CARGO is publicly available at \texttt{GitHub} repository~\cite{CARGO2025}.
% $\gtrapprox$ 5\% increase in successful deliveries and a more than 60\% reduction in charging costs compared to the baselines.}
% The framework is evaluated using two datasets: (1) real-world data extracted from the \texttt{Google Maps} API for \texttt{St. Louis, Missouri}, and (2) a synthetic dataset generated by selecting a depot in \texttt{San Francisco} and applying the \texttt{Haversine} formula to simulate delivery and charging point locations within a $30$-\textit{mile} radius.
% \item The framework is evaluated using two datasets: (1) real-world data extracted from the \texttt{Google Maps} API for St. Louis, Missouri, and (2) a synthetic dataset generated by selecting a depot in San Francisco and using the Haversine function to simulate delivery and charging point locations within a $30$-\textit{mile} radius. Performance is assessed in two phases: first, by comparing the optimal solution (OPT), implemented with \texttt{Gurobi} on an HPC cluster, against the heuristic; second, by benchmarking the heuristic against the Earliest Deadline First (EDF) and Nearest Delivery First (NDF) heuristics, implemented in \texttt{Python} on a local machine. Results demonstrate a $\gtrapprox$5\% increase in successful deliveries and a $\gtrapprox$60\% reduction in charging costs compared to baseline methods, which is substantial. 
\end{enumerate}

The paper is organized as follows: Section \ref{sec:related_work} reviews related literature, while Section \ref{sec:system_model} details the system model. Scheduling constraints are outlined in Section \ref{sec:constraints_on_scheduling}, followed by the EDRP problem formulation in Section \ref{sec:MILP_Formulation}. Section \ref{sec:solution_approach} describes the proposed methodology. Experimental setup and results are presented in Section \ref{sec:result}, and Section \ref{sec:cnls} concludes the paper.
% \vspace{-0.2in}
% \vspace{-0.1in}
\section{Related Work}\label{sec:related_work} 
% \vspace{-0.05in}
The reviewed literature falls into two categories: (1) delivery optimization, focusing on routing vehicles to meet delivery time windows, and (2) EV charging strategies, which assign charging points (CPs) based on vehicle SoC and expected wait times.
\begin{table}[t]
\caption{Comparative Analysis of Reviewed Literature.}
\label{tab:comparitive_study}
\centering
\small
\resizebox{\columnwidth}{!}{%
\begin{tabular}{|p{0.7cm}|p{5.6cm}|c|c|c|c|}
\hline
\textbf{Refs.} & \textbf{Solution Approach} & \rotatebox{90}{Route Opt.} & \rotatebox{90}{Time Window} & \rotatebox{90}{Charging Strat.} & \rotatebox{90}{Real-Time Data} \\ \hline
\cite{suh2012leveraging} & Alternative pickup locations & \cmark & \xmark & \xmark & \xmark \\ \hline
\cite{han2017appointment} & Approximate scheduling with routing & \cmark & \cmark & \xmark & \xmark \\ \hline
\cite{zhou2018multi} & Two-stage genetic algorithm & \cmark & \xmark & \xmark & \xmark \\ \hline
\cite{moshref2020truck} & Multi-modal (trucks and UAVs) & \cmark & \xmark & \xmark & \xmark \\ \hline
\cite{boysen2022crowdshipping} & Incentive-driven crowd shipping & \cmark & \xmark & \xmark & \xmark \\ \hline
\cite{fan2022food} & Evolutionary game theory analysis & \xmark & \xmark & \xmark & \xmark \\ \hline
\cite{Lastmile2023} & Non-cooperative game for EV routing & \cmark & \cmark & \cmark & \xmark \\ \hline
\cite{algafri2024smart} & AHP w/ goal programming for EV charging & \xmark & \xmark & \cmark & \xmark \\ \hline
\cite{sassi2017electric} & EV scheduling model & \xmark & \xmark & \cmark & \xmark \\ \hline
\cite{elghitani2020efficient} & Crowd detection w/ Benders decomposition & \xmark & \xmark & \cmark & \cmark \\ \hline
\cite{tao2023distributed} & Lyapunov optimization for EV assignment & \xmark & \xmark & \cmark & \cmark \\ \hline
\cite{bus-coop} & MILP for electric bus fleets & \cmark & \cmark & \cmark & \xmark \\ \hline
\cite{danchuk2021simulation} & ACO with traffic-aware routing & \cmark & \xmark & \xmark & \cmark \\ \hline
\cite{danchuk2023optimization} & ACO-based dynamic routing & \cmark & \xmark & \xmark & \cmark \\ \hline
\cite{alvarez2024optimizing} & Multi-objective freight routing & \cmark & \xmark & \xmark & \xmark \\ \hline
\cite{khanda2025smevca} & EV-CP assignment using stable matching & \xmark &\xmark &\cmark &\xmark \\ \hline
\textbf{EDRP} & MILP w/ heuristic for EV delivery routing & \cmark & \cmark & \cmark & \cmark \\ \hline
% \hline
\end{tabular}
}
% \vspace{-0.1in}
\end{table}

\noindent \textbf{Delivery Optimization:} The authors in \cite{suh2012leveraging} explored social networks for package deliveries, proposing pickup locations like kiosks at transit stations and grocery stores to reduce emissions. While effective in urban settings, this approach may be less suitable for suburban or rural areas, which could result in increased emissions. Addressing scheduling, \cite{han2017appointment} integrated vehicle routing with appointment scheduling under soft time windows, employing dynamic programming (DP) and tabu search. Building on cost efficiency, \cite{zhou2018multi} proposed a two-stage genetic algorithm (GA)-based solution that transports packages to satellite locations before final delivery. 
Alternatively, \cite{danchuk2021simulation, danchuk2023optimization}   propose ant colony optimization (ACO)-based approaches for dynamic cargo delivery routing in urban environments. While the former integrates ACO with a synergetic traffic flow model to account for non-stationary dynamics, the latter enhances real-time route updates by utilizing IoT-based traffic data and historical data within the smart logistics framework.
On the other hand, the work in \cite{alvarez2024optimizing} proposed a route optimization approach that simultaneously minimizes travel time and distance by translating both into monetary terms using vehicle operating costs and the value of time. This cost-based formulation shows improvement over traditional distance-only methods, particularly in congested urban areas, and enables more informed decision-making in logistics operations.

To minimize waiting times, \cite{moshref2020truck} and \cite{khanda2022drone} introduced a multimodal framework combining trucks and UAVs. For crowdsourced delivery, \cite{boysen2022crowdshipping} developed an incentive-driven framework where distribution center employees deliver online orders on their way home, using Benders decomposition to maximize matched shipments while ensuring minimum earnings. Addressing traffic safety, \cite{fan2022food} employed evolutionary game theory to analyze the decision-making behaviors of delivery drivers and food delivery platforms. Focusing on EV integration, \cite{Lastmile2023} introduced a routing solution for crowdsourced EV drivers on pre-planned trips, modeling multi-destination deliveries within time windows as a non-cooperative game, solved using stable matching and trimmed DP.

\noindent \textbf{EV Charging Strategies:}
Several studies have optimized EV assignments to CSs. In \cite{algafri2024smart}, the Analytic Hierarchy Process, combined with goal programming, minimized costs, battery degradation, and time overheads for EV owners. However, their approach overlooks extended trips that require multiple charging stops, a challenge addressed in \cite{sassi2017electric}, which proposed a scheduling model that maximizes travel distance while minimizing costs, without considering real-time charging availability. In ride-sharing services, \cite{elghitani2020efficient} optimized EV-to-charging assignments using crowd detection and generalized Benders decomposition, modeling CSss as $M/M/D$ queues and employing a distributed biased min-consensus algorithm for navigation. To reduce complexity, \cite{tao2023distributed} proposed a Lyapunov optimization-based approach to minimize total service time by dynamically reassigning EVs based on travel delays; however, frequent reassignments could disrupt travel and lead to dissatisfaction. Alternatively,~\cite{bus-coop} co-optimized route planning, charging locations, and scheduling for electric bus fleets using MILP, and iterated local search for effective decision-making. \cite{khanda2025smevca} discusses a stable, fast, and efficient EV-CP assignment framework, formulating it as a one-to-many matching game with preferences. The authors devised two coalition formation strategies, namely, the preferred coalition greedy (PCG) and the preferred coalition dynamic (PCD), to compute the suboptimal and optimal coalitions, respectively.

\noindent \textbf{Research Gaps:}
As summarized in Table~\ref{tab:comparitive_study}, existing research on EV routing~\cite{Lastmile2023, bus-coop, danchuk2021simulation, danchuk2023optimization, alvarez2024optimizing}, charging strategies~\cite{algafri2024smart, elghitani2020efficient, tao2023distributed, khanda2025smevca}, and delivery logistics~\cite{suh2012leveraging, han2017appointment, zhou2018multi, boysen2022crowdshipping, fan2022food} often addresses these components in isolation. While many approaches optimize EV assignment and scheduling, they often overlook real-time energy constraints, battery limitations, and charging costs, leading to suboptimal routing decisions. Additionally, multi-stop charging is frequently overlooked. The proposed \textsc{CARGO} framework bridges these gaps by jointly optimizing EV routing and charging, incorporating real-time charging availability and cost-awareness while considering delivery windows to enhance operational efficiency. Finally, Table~\ref{tab:comparitive_study} highlights key features of related works in EV routing and logistics, comparing solution approaches based on their support for route optimization, time window constraints, charging strategies, and real-time data integration.

% However, with the inevitable adoption of EVs for last mile delivery, has not been ignored in these works, particularly regarding charging considerations, which directly affect operational costs, delivery times, and customer satisfaction.
% \vspace{-0.07in}
\section{System Model and Assumptions}\label{sec:system_model}
% \vspace{-0.1in}
% \begin{figure}
%     \centering
%     \includegraphics[width=\linewidth]{Tex/Figures/EDT-2.drawio.png}
%     \caption{EV-based Delivery Planning }
%     \label{fig:SysModel}
% \end{figure}
This study presents a  delivery framework CARGO, as illustrated in Fig. \ref{fig:IntroModel}, with multiple EVs operating from a distribution center. EVs complete deliveries within set time windows and return, recharging en route based on cost, proximity, and SoC. The problem involves delivery and charge scheduling, and the system model is detailed next.
% We outline the system model, covering (1) delivery windows, (2) recharging operations, and (3) assumptions, followed by constraints and MILP formulation.
% Consider a scenario where goods ordered from an online retailing company are delivered to the nearest distribution center in the delivery city. From this distribution center, goods must be distributed to their final delivery point in the user-specified time window.
% Figure \ref{fig:SysModel} depicts this last-mile delivery scenario with the distribution center managing multiple electric delivery trucks.
% Each delivery truck starts at the distribution center every morning, is assigned a load of goods to be delivered within their time windows, and returns to the distribution center at the end of the day. 
% We propose a novel cost-optimal approach to plan the delivery trip for the available electric trucks while ensuring low-cost charging and minimal energy consumption for every truck with the maximum number of deliveries catered successfully. 
%\vspace{-0.05in}

% \vspace{2pt}
\noindent \textbf{System Model: } We consider a delivery system with a single distribution center ($d_0$) and delivery points $\mathcal{D} = \{d_1,$ $d_2,$ $ \cdots,$ $d_i,$ $\cdots, d_{|\mathcal{D}|}\}$. 
% Figure \ref{fig:SysModel} depicts the system model. 
% To deliver the items in $\mathcal{D}$, we have a set of EV delivery trucks $\mathcal{E} =\{e_1, e_2, \cdots, e_j \cdots, e_{|\mathcal{E}|}\}$ that start at the distribution center $d_0$ and carry out the deliveries en-route and return to the distribution center. 
A fleet of EVs, represented as $\mathcal{E} = \{e_1, e_2, \ldots, e_j, \ldots, e_{|\mathcal{E}|}\}$, is utilized for item deliveries. Each EV $e_j \in \mathcal{E}$ embarks on its journey from the distribution center $d_0$, completes deliveries along its designated route, recharges if necessary, and returns to $d_0$. An EV can deliver multiple items in a single trip, provided it adheres to the specified time window associated with each delivery.

% \vspace{2pt}
\noindent \textbf{Delivery Window:} Each delivery $d_i \in \mathcal{D}$ is associated with a time window $\left[\tau_i^{start}, \tau_i^{end}\right]$, where $\tau_i^{start}$ is the earliest allowable delivery time and $\tau_i^{end}$ is the latest allowable delivery time. For example, if $e_j$ delivers $d_i$ and is then scheduled to deliver $d_{i'}$, $t_{i^{'}.dep} \leq \tau_{i'}^{end}$ captures the successful delivery of $d_{i'}$. Note that $t_{i^{'}\cdot arr} = t_{i\cdot dep} + \gamma(i, i')$, where $t_{i\cdot dep}$ is the departure time of $e_j$ from $d_i$ after completing the delivery, and $\gamma(i, i')$ is the traversal time from $d_i$ to $d_{i'}$.
%\vspace{-0.1in}
% \begin{equation}\label{eqn:cons1}
% % \vspace{-0.1in}
%     % t_{i'}^{j, arr} \leq \tau_{i'}^{end} 
%     t_{i^{'}.dep} \leq \tau_{i'}^{end} 
% % \vspace{-0.1in}
% \end{equation}
% Additionally, there is a possibility that 
\begin{table}[t]
\centering
\footnotesize
\caption{Notations Used.}
\label{tab:tab_notations}
 \resizebox{\columnwidth}{!}{
\begin{tabular}{|c|l|}
\hline  
{\bf Notation} & {\bf Description} \\ \hline \hline
$\mathcal{D}$ & The set of deliveries. \\ \hline
$d_0$ & The depot or distribution center. \\ \hline
$\mathcal{E}$ & The set of EVs. \\ \hline
$e_j$ & The $j^{th}$ EV. \\ \hline
$d_i$ & The $i^{th}$ delivery.  \\ \hline
$\tau_i^{start}$ & Earliest delivery time for the $i^{th}$ delivery.  \\ \hline
$\tau_i^{end}$ & The latest delivery time for the $i^{th}$ delivery. \\ \hline
$t_{i\cdot arr}$ &  The arrival time at delivery point $d_{i}$. \\ \hline
$t_{i\cdot dep}$ & The departure time at delivery point $d_i$. \\ \hline
$\hat{t}^{j, l}_{y \cdot arr}$ & The arrival time of $e_j$ at CP $y$ on its $l^{th}$ subtrip. \\ \hline
$\hat{t}^{j, l}_{y \cdot dep}$ & The departure time of $e_j$ at CP $y$ on its $l^{th}$ subtrip. \\ \hline
% $\hat{t}_{i, j, l}^{arr}$ & The arrival time of EV $e_j$ at charging point $d_i$ on its $l^{th}$ subtrip. \\\hline
% $\hat{t}_{i, j, l}^{dep}$ & The departure time of EV $e_j$ at charging point $d_i$ on its $l^{th}$ subtrip. \\\hline
$\gamma(x, y)$[\textit{min}] & Time taken to traverse from points $x$ to $y$. \\ \hline
$\Gamma_{un} [\textit{min}]$ & Package unloading time. \\ \hline
$\delta_{i, i'}$[\textit{miles}] & The distance between the delivery points $d_i$ and $d_{i'}$. \\ \hline
$\mathcal{S}$ & Set of all subtrips \\ \hline
$s_l$ & A specific subtrip $l$ \\ \hline
$a_j [\textit{miles/min}]$ & The average velocity of the vehicle $e_j$ in $s_l$. \\ \hline
$\beta^f$ [\textit{kwh}] & The battery capacity of a EV in $\mathcal{E}$. \\ \hline
$\beta^{j, l}$ [\textit{kwh}] & The charge needed by $j^{th}$ for the subtrip $s_l$. \\ \hline
$\Psi(x, y)$ & The energy expended in traversing from points $x$ to $y$. \\ \hline
$u_x^{j, l}$ & The sequence number of visitation for $x \in \mathcal{D}$ by $e_j$. \\  \hline
% $u_x^{j, l}$ & The sequence number of visitation for $x \in \mathcal{D}$ by $e_j$ \\ \hline
$m_j$ & The mileage of the EV $e_j$. \\ \hline 
$\mathcal{C}$ & Set of all CPs. \\ \hline 
$c_q$ & A unique CP $q$. \\ \hline 
$r_j$ & Charge acceptance rate of the vehicle $e_j$. \\ \hline
$\hat{r}_y$ & Charging rate of CP. \\ \hline
$\theta_y$[$\$/kwh$] & The energy cost for charging at the $y$. \\ \hline
$W_y$ & Average waiting time at the CP $y$. \\ \hline
$\alpha_1, \alpha_2$ & Normalization Factors. \\ \hline
\end{tabular}
}
% \vspace{-0.05in}
\end{table}
If $e_j$ arrives at $d_i$ before the start of the delivery window, i.e., $t_{i \cdot arr} < \tau_i^{start}$, the delivery cannot be completed before $\tau_i^{start}$. In such cases, $e_j$ waits for the duration $\tau_i^{start} - t_{i \cdot arr}$. 
The item can be delivered once the window opens, and $e_j$ departs after delivery.
% , i.e., $t_{i \cdot dep} = \tau_i^{start}$. 
% \textcolor{red}{For all other deliveries, the departure time is the same as the arrival time, i.e., $t_{i \cdot dep} = t_{i \cdot arr}$.} 
The delivery of any $d_i \in \mathcal{D}$ must occur within the specified time window, i.e., $\tau_i^{start} \leq t_{i \cdot dep} \leq \tau_i^{end}$. 
The proposed scheme can easily be modified to include a delivery deadline instead of time windows by setting $\tau_i^{start} = 0$.
% as captured in Eq. (\ref{eqn:departure_time_window}).
% \vspace{-0.1in}
% \begin{equation}\label{eqn:departure_time_window}
% % \vspace{-0.1in}
%    \tau_i^{start} \leq t_{i \cdot dep} \leq \tau_i^{end}
% \end{equation}
Additionally, a fixed amount of time spent on unloading and delivering the item is captured by $\Gamma_{un}$ [\textit{min}] \cite{bus-coop}, and consequently the departure time of $e_j$ after delivering $d_i$ can be expressed as $t_{i \cdot dep} = t_{i \cdot arr} + \Gamma_{un}$.
% \textcolor{blue}{Additionally, we also maintain an indicator variable $y_d \in [0, 1]$, is set to $1$ for successful delivery, otherwise $0$.}

% \vspace{2pt}
\noindent \textbf{Delivery Time and Distance:} For any two successive deliveries, $d_i$ and $d_{i'}$, we assume that the designated EV takes the shortest route $\delta_{i, i'}$ [\textit{miles}]. Given the average velocity $a_j$ [\textit{miles/min}] of $e_j$, the traversal time is calculated as $\gamma(i, i')$ = $\delta_{i, i'}/a_j$ [\textit{min}]. 
Here, we assume consistent traffic conditions, as our primary objective is to charge and route to maximize successful deliveries co-optimally. However, our model can be readily extended to incorporate complex traffic models such as \cite{greenshields1935study}, enabling more accurate EV arrival predictions.

% \vspace{2pt}
\noindent \textbf{Recharging the EVs:} The battery of an EV is limited and may require recharging during a trip to complete all deliveries. All the EVs in $\mathcal{E}$ are assumed to have a uniform battery capacity of $\beta^f$ [\textit{kWh}]. To determine the SoC of an EV, we define the notion of a \textit{subtrip}, which represents a sequence of deliveries between two charging sessions. In a \textit{subtrip}, the starting and ending points are charging locations, while the intermediate points are delivery locations. Low-cost charging can be done overnight at the distribution center before the trip or at intermediate public CSss at a higher cost during the trip. 
This is necessary because every charging location, including the distribution center, has a location-variant price, denoted by $\theta_q, \forall c_q \in \mathcal{C} \cup \{d_0\}$, where $\mathcal{C}$ is the set of CPs. A failed delivery is returned to the distribution center upon trip completion. The set of \textit{subtrips} is denoted by $\mathcal{S} = \{s_1, s_2, \cdots, s_l, \cdots, s_{|\mathcal{S}|}\}$, where $s_l$ uniquely identifies the $l^{th}$ \textit{subtrip}. The rest of the notations are captured in Table \ref{tab:tab_notations}. 
% For instance, let $s = \{c_k, d_1, d_2, \cdots, c_{k'} \}$ be a subtrip, wherein the start and the end-points of the subtrip are charging operations done at charging points $c_k, c_{k'} \in \mathcal{C} \cup \{\mathcal{D}_0\}$. Note that $\mathcal{C}$ is the set of all CSs considered, and all other points $d_{i} \in s \setminus \{c_k, c_{k'}\}$ are delivery points such that $ 1 \leq i \leq n$. 
%Additionally, we assume that at every charging session, i.e., (1) low-cost overnight charging at the distribution center ($d_0$) or (2) intermediate charging at the public CS, an EV comes out fully charged \cite{?}.
% \vspace{0.08in}

\noindent \textbf{Binary Indicator Variables}: We define the following indicator variables. Eq. (\ref{eqn:indicator}) is set to $1$ if $e_j$ traverses the point $x$ followed by $y$ in the subtrip $s_l$, wherein $x, y$ can be delivery points or CPs or the distribution center itself. 

% \begin{small}
\vspace{-0.1in}
\begin{align}
\label{eqn:indicator}
\rchi_{x, y}^{j, l} = \left\{
            \begin{array}{ll}
            1 & \text{$e_j$ travels from $x$ to }  \\
            & \text{$y$ in the sub-trip $s_l$.} \\
            0 & \text{otherwise.}
            \end{array}
\right.
\end{align}
\vspace{-0.1in}
% \end{small}

We validate a sub-trip by the indicator variable defined as per Eq. (\ref{eqn:indicator_2}). It states that a sub-trip is valid if at least one delivery point exists after charging.

% \begin{small}
\vspace{-0.1in}
\begin{align}
\label{eqn:indicator_2}
\mathcal{Z}^{j, l} = \left\{
            \begin{array}{ll}
            1 & \text{If the sub-trip $s_l$ is assigned} \\  & \text{to vehicle $e_j$} \\
            0 & \text{otherwise.}
            \end{array}
\right.
\end{align}
\vspace{-0.1in}

\noindent \textbf{Assumptions:} EVs commence their routes fully charged from the distribution center, recharging at designated stations as necessary. 
Moreover, in CARGO, we disallow consecutive charging sessions without a delivery, ensuring at least one delivery occurs between charges. 
During recharging, EVs are fully charged to 100\%; however, with minor adjustments, the CARGO framework can accommodate partial recharge scenarios. The distribution center offers the most economical charging rates, making it ideal for overnight charging. Although charging costs differ by location, they remain constant at each station throughout the day
\section{Constraints}\label{sec:constraints_on_scheduling}
% \vspace{-0.05in}
The delivery model presented in this paper operates under constraints detailed in Tables~\ref{tab:table_1} to \ref{tab:table_4}, which reflect real-world conditions. For clarity, these constraints are categorized based on different aspects of the delivery process. This structured approach ensures that each constraint is systematically addressed, facilitating a comprehensive model analysis.
\begin{table}[!htbp]
    \centering
    \normalsize
    % \vspace{-0.1in}
    \caption{\vspace{5pt} \normalsize Flow Constraints.}
    \label{tab:table_1}
    \small
    % \rowcolors{2}{SeaGreen!10}{White}
    \rowcolors{2}{White}{White}
    % \hspace{-0.2in}
    % \begin{minipage}{0.48\textwidth}
        \centering
        % \begin{tabular}{|>{\columncolor{Emerald}}p{0.06in}|>{\columncolor{Emerald}}p{3.1in}|} 
        \renewcommand{\arraystretch}{1.25}
         \vspace{-0.1in}
        \begin{tabular}{|>{\columncolor{Emerald}}>{\raggedleft\arraybackslash}p{0.13in}|>{\columncolor{Emerald}}p{3.05in}|} 
            \hline
            {\color{white}\bf ID} & {\color{white}\bf Flow Constraints} \\ \hline \hline
            C1   & $\sum_{y \in \mathcal{D}} \rchi_{d_0, y}^{j, l} \leq 1, \hfill\,\, \forall e_j \in \mathcal{E}, l = 0$ \\ \hline
            C2   & $\sum_{s_l \in S}\sum_{y \in \mathcal{D}} \rchi_{d_0, y}^{j, l} 
            = \sum_{s_l \in \mathcal{S}}  \sum_{x \in \mathcal{D}} \rchi_{x, d_0}^{j, l} \leq 1,  \, \, \forall e_j \in \mathcal{E}$ \\ \hline
            C3 &  
            $\sum_{x \in \mathcal{D}}  \rchi_{x, y}^{j, l} 
            = \sum_{x \in \mathcal{D}} \rchi_{y,x}^{j, (l+1)},$  $ \hfill
            \,\, \forall e_j \in \mathcal{E}, \forall s_l \in \mathcal{S}, \forall y \in \mathcal{C}$ \\ \hline
            % $\sum_{x \in \mathcal{D}} \sum_{y \in \mathcal{C}} \rchi_{x, y}^{j, l} 
            % = \sum_{x \in \mathcal{C}} \sum_{y \in \mathcal{D}} \rchi_{x, y}^{j, (l+1)},$ \\ & $ \hfill
            % \,\, \forall e_j \in \mathcal{E}, \forall s_l \in \mathcal{S}$ \\ \hline
            C4 & $( \sum_{y \in \mathcal{C}} \rchi_{y, x}^{j, l} + \sum_{y \in \mathcal{D} \setminus \{x\}} \rchi_{y, x}^{j, l} + \rchi_{d_0, x}^{j, l}) $\\ & $\quad \quad \quad = ( \sum_{y \in \mathcal{C}} \rchi_{x, y}^{j, l} + \sum_{y \in \mathcal{D} \setminus \{x\}} \rchi_{x, y}^{j, l} + \rchi_{x, d_0}^{j, l}),$ \\ & $\hfill \,\, \forall e_j \in \mathcal{E}; \forall s_l \in \mathcal{S}; \forall x \in \mathcal{D}$ \\ \hline
            C5 & $\sum_{s_l \in \mathcal{S}} \sum_{e_j \in \mathcal{E}} (\sum_{x \in \mathcal{D}, \, x \neq y} \rchi_{x, y}^{j, l} + \sum_{x \in \mathcal{C} \cup \{d_0\}} \rchi_{x, y}^{j, l} ) \leq 1,$ $\hfill \, \forall y \in \mathcal{D}$ \\ \hline
        \end{tabular}
        \vspace{-0.1in}
\end{table}
\\
\textbf{Flow Constraints:} To ensure a consistent flow of EVs across all nodes—depots, delivery locations, and CPs—the following constraints are implemented: Constraint C1 stipulates that each EV departs from the distribution center no more than once; Constraint C2 mandates that an EV returns to the distribution center only after completing deliveries over one or more sub-trips; Constraint C3 maintains CP flow consistency by balancing the inflow and outflow of EVs; Constraint C4 enforces similar flow consistency at delivery points; and Constraint C5 ensures that a delivery point can be visited at most once. Constraints C4 and C5 also specify that a delivery location can be reached from a CP, a previous delivery location, or directly from the distribution center if it is the first delivery.
\begin{table}[!htbp]
    \centering
    \normalsize
    \vspace{-0.07in}
    \caption{\vspace{5pt} \normalsize Subtrip Constraints.}
    \label{tab:table_2}
    \small
    % \rowcolors{2}{SeaGreen!10}{White}
    \rowcolors{2}{White}{White}
    % \hspace{-0.2in}
    % \begin{minipage}{0.48\textwidth}
        \centering
        % \begin{tabular}{|>{\columncolor{Emerald}}p{0.06in}|>{\columncolor{Emerald}}p{3.1in}|} 
        \renewcommand{\arraystretch}{1.25}
        \vspace{-0.1in}
        \begin{tabular}{|>{\columncolor{Emerald}}>{\raggedleft\arraybackslash}p{0.18in}|>{\columncolor{Emerald}}p{3.05in}|} 
            \hline
            {\color{white}\bf ID} & {\color{white}\bf Subtrip Constraints} \\ \hline \hline
            C6 & $\sum_{x \in  \mathcal{C} \cup \{d_0\}} \sum_{y \in \mathcal{D}} \rchi_{x, y}^{j, l} - M \cdot \mathcal{Z}^{j, l}
            \leq 0, \quad \forall e_j \in \mathcal{E}, \forall s_l \in \mathcal{S}$ \\ \hline
            C7 & $\sum_{x \in  \mathcal{C} \cup \{d_0\}} \sum_{y \in \mathcal{D}} \rchi_{x, y}^{j, l}  + M(1- \mathcal{Z}^{j, l}) \geq 0 ,\,\,$\\ & $\hfill \forall e_j \in \mathcal{E}, \forall s_l \in \mathcal{S}$ \\ \hline
            C8 & $\sum_{x \in  \mathcal{C} \cup \{d_0\}} \sum_{y \in \mathcal{D}} \rchi_{x, y}^{j, l} = \mathcal{Z}^{j, l}, \hfill \forall e_j \in \mathcal{E}, \forall s_l \in \mathcal{S}$ \\ \hline
            C9 & $\sum_{x \in \mathcal{D}} \sum_{y \in  \mathcal{C} \cup \{d_0\}}\rchi_{x, y}^{j, l} = \mathcal{Z}^{j, l}, \hfill \forall e_j \in \mathcal{E}, \forall s_l \in \mathcal{S}$ \\ \hline
            C10 & $\sum_{x \in \mathcal{D}} \rchi_{x, z}^{j, l}  - \sum_{y \in \mathcal{D}} \rchi_{z, y}^{j, l+1} = 0, 
            $\\ & $\hfill \forall s_l \in \mathcal{S} , \,\, \forall e_j \in \mathcal{E}, \,\,  \forall z \in \mathcal{C}$ \\ \hline
            C11 &  $u_x^{j,l} - u_y^{j,l} + n * \rchi_{x, y}^{j, l} \leq n-1 , \,\, \forall e_j \in \mathcal{E}; \, \forall s_l \in \mathcal{S},\, \forall x, y \in \mathcal{D}$ \\ \hline
            C12 & $2 \leq u_{x}^{j, l} \leq n; \hfill \forall e_j \in \mathcal{E}; \, \forall s_l \in \mathcal{S}, \,\, \forall x \in \mathcal{D}$ \\ \hline
        \end{tabular}
         \vspace{-0.05in}
\end{table}
% \vspace{-0.1in}

\noindent\textbf{Subtrip Constraints:}
These constraints ensure the proper visit sequence of an EV between two charging sessions. Constraints C6 and C7 validate a sub-trip by ensuring that the EV visits at least one delivery location between consecutive charging sessions. This is modeled using the \texttt{Big-M} method for accurate conditional representation \cite{wolsey1999integer}, where $M$ is a large constant. 
Constraints C8 and C9 enforce that each sub-trip starts and ends at a CP. The sequence number of visiting $x \in \mathcal{D}$ by $e_j$ in sub-trip $s_l$ is represented as $u_x^{j, l}$. The order in which an EV traverses sub-trips follows a structured sequence captured in Constraint C10, adhering to the \texttt{Miller-Tucker-Zemlin (MTZ)} sub-tour elimination criteria \cite{MTZ-TSP-91}. Constraints C11 and C12 enforce the \texttt{MTZ} conditions, where C11 eliminates sub-tours, and C12 defines the valid range of values for $u_x^{j,l}$. Although $u_x^{j,l}$ is a continuous decision variable, C11 and C12 remain linear since they do not involve product terms of decision variables.
\begin{table}[!htbp]
    \centering
    \normalsize
    % \vspace{-0.1in}
    \caption{\vspace{5pt} \normalsize Charging Constraints.}
    \label{tab:table_3}
    \small
    % \rowcolors{2}{SeaGreen!10}{White}
    \rowcolors{2}{White}{White}
    % \hspace{-0.2in}
    % \begin{minipage}{0.48\textwidth}
        \centering
        % \begin{tabular}{|>{\columncolor{Emerald}}p{0.06in}|>{\columncolor{Emerald}}p{3.1in}|} 
        \renewcommand{\arraystretch}{1.25}
         \vspace{-0.1in}
        \begin{tabular}{|>{\columncolor{Emerald}}>{\raggedleft\arraybackslash}p{0.18in}|>{\columncolor{Emerald}}p{3.05in}|} 
            \hline
            {\color{white}\bf ID} & {\color{white}\bf Charging Constraints} \\ \hline \hline
            C13 & $\beta^{j,l} = \sum_{y \in \mathcal{C} \cup \{d_0\}}\sum_{x \in \mathcal{D}} \rchi_{y, x}^{j, l} \cdot \Psi(y, x) + \sum_{x \in \mathcal{D}}\sum_{y \in \mathcal{D}\setminus \{x\}} \rchi_{x, y}^{j, l} \cdot \Psi(x, y) + \sum_{x \in \mathcal{D}} \sum_{y \in \mathcal{C} \cup \{d_0\}} \rchi_{x, y}^{j, l} \cdot \Psi(x, y), \qquad \forall s_l \in \mathcal{S}, \hfill \forall e_j \in \mathcal{E}$ \\ \hline
            C14 & $\beta^{j,l} \leq  \beta^f; \,\, \forall s_l \in \mathcal{S}, \hfill\,\, \forall e_j \in \mathcal{E}$     \\ \hline
 \end{tabular}
  % \vspace{-0.1in}
\end{table}
\noindent \textbf{Charging Constraints:}
Constraint C13 calculates the total energy expended in a sub-trip, encompassing (1) the traversal from a CP (or the distribution center, if it is the first delivery) to the first delivery point in the sub-trip; (2) intermediate deliveries between charging sessions; and (3) the final delivery before reaching the next CS or returning to the distribution center. Here, $\beta^{j,l}$ is a continuous decision variable that captures the total charge needed to complete a subtrip. 
The energy expended in traveling from point $x$ to point $y$, represented as $\Psi(x, y)$, is computed as the ratio of the shortest distance $\delta_{x, y}$ [\textit{miles}] to the vehicle's average mileage $m_j$ [\textit{miles/kWh}]. 
Constraints C13 remains linear due to the absence of product terms among decision variables, whereas C14 ensures that an EV completes all its deliveries without fully depleting its battery that means $\beta^{j,l}$ is bounded by $\beta^{f}$. 
% \begin{subequations}\label{eqn:trip_battery_constraint}
% \small
% \vspace{-0.15in}
% \begin{align*}
% % \begin{split}
% \beta^{j,l} =
% & \sum_{y \in \mathcal{C} \cup \{d_0\}}\sum_{x \in \mathcal{D}} \rchi_{y, x}^{j, l} \cdot \Psi(y, x) + \sum_{x \in \mathcal{D}}\sum_{y \in \mathcal{D}\setminus \{x\}} \rchi_{x, y}^{j, l} \cdot \Psi(x, y) \notag 
% % \label{eqn:trip_battery_constraintpart1} 
% \\ 
% & + \sum_{x \in \mathcal{D}} \sum_{y \in \mathcal{C} \cup \{d_0\}} \rchi_{x, y}^{j, l} \cdot \Psi(x, y),
% % \leq \beta^f; 
% \qquad \forall s_l \in \mathcal{S}, 
% \forall e_j \in \mathcal{E} 
% \\
% & \qquad \qquad  \beta^{j,l} \leq  \beta^f; \,\, \forall s_l \in \mathcal{S}, \,\, \forall e_j \in \mathcal{E}   
% % \label{eqn:trip_battery_constraintpart2}
% % \end{split}
% \end{align*}
% \vspace{-0.1in}
% \end{subequations}
% \vspace{-0.05in}
\begin{table}[!htbp]
    \centering
    \normalsize
    \vspace{-0.05in}
    \caption{\vspace{5pt} \normalsize Time Constraints.}
    \label{tab:table_4}
    \small
    % \rowcolors{2}{SeaGreen!10}{White}
    \rowcolors{2}{White}{White}
    % \hspace{-0.2in}
    % \begin{minipage}{0.48\textwidth}
        \centering
        % \begin{tabular}{|>{\columncolor{Emerald}}p{0.06in}|>{\columncolor{Emerald}}p{3.1in}|} 
        \renewcommand{\arraystretch}{1.25}
         \vspace{-0.1in}
        \begin{tabular}{|>{\columncolor{Emerald}}>{\raggedleft\arraybackslash}p{0.18in}|>{\columncolor{Emerald}}p{3.05in}|} 
            \hline
            {\color{white}\bf ID} & {\color{white}\bf Time Constraints} \\ \hline \hline
            C15 & $\hat{t}^{j, l}_{d_0\cdot dep} \geq 0, \,\,\, \forall e_j \in \mathcal{E}, l = 1$ \\ \hline 
            % C16 & $t_{y \cdot arr} \geq  \sum_{x \in \mathcal{D}} \rchi_{x, y}^{j,l}(t_{x \cdot. dep} +  \gamma(x, y)) + \sum_{x \in \mathcal{C}} \rchi_{x, y}^{j, l}(\hat{t}^{j, l}_{x\cdot dep} +  \gamma(x, y)), \hfill \forall e_j \in \mathcal{E}, \forall s_l \in \mathcal{S}, \forall y \in \mathcal{D}$ \\ \hline
            C16 &   $\Omega_{x,y}^{j,l} \geq 0; $ $\,\,\, \Omega_{x,y}^{j,l} \leq t_{x \cdot dep};$ $\,\,\, \Omega_{x,y}^{j,l} \leq M \cdot \rchi_{x,y}^{j,l};$\\ &
            $\Omega_{x,y}^{j,l} \geq t_{x \cdot dep} - M (1 - \rchi_{x,y}^{j,l});$ $\hfill \forall e_j \in \mathcal{E}, \forall s_l \in \mathcal{S}, \forall x, y \in \mathcal{D}$\\ \hline
            C17 &   $\hat{\Omega}_{x,y}^{j,l} \geq 0; $ $\,\,\, \hat{\Omega}_{x,y}^{j,l} \leq \hat{t}_{x \cdot dep}^{j,l};$ $\,\,\, \hat{\Omega}_{x,y}^{j,l} \leq M \cdot \rchi_{x,y}^{j,l};$\\ &
            $\hat{\Omega}_{x,y}^{j,l} \geq \hat{t}_{x \cdot dep}^{j,l} - M (1 - \rchi_{x,y}^{j,l});$ \\ & $\hfill \forall e_j \in \mathcal{E}, \forall s_l \in \mathcal{S}, \forall x \in \mathcal{C}, \forall y \in \mathcal{D}$\\ \hline
            C18 & $t_{y \cdot arr} \geq  \sum_{x \in \mathcal{D}} (\Omega_{x,y}^{j,l} +  \rchi_{x, y}^{j,l} \cdot \gamma(x, y)) + \sum_{x \in \mathcal{C}} (\hat{\Omega}_{x,y}^{j,l} +  \rchi_{x, y}^{j, l} \cdot \gamma(x, y)), \hfill \forall e_j \in \mathcal{E}, \forall s_l \in \mathcal{S}, \forall y \in \mathcal{D}$ \\ \hline
            C19 & $\hat{t}^{j, l}_{y \cdot arr} \geq \sum_{x \in \mathcal{D}} (\Omega_{x,y}^{j,l} +  \rchi_{x, y}^{j,l} \cdot \gamma(x, y)),
             \forall e_j \in \mathcal{E}, \forall s_l \in \mathcal{S}, \forall y \in \mathcal{C}$ \\ \hline
            C20 & $t_{y \cdot dep} \geq t_{y \cdot arr},  \hfill \forall y \in \mathcal{D}
            $\\ & $ \tau_{y}^{start} + \Gamma_{un} \leq t_{y \cdot dep} \leq \tau_{y}^{end}, \hfill \forall y \in \mathcal{D}$ \\ \hline
            C21 & $\hat{t}^{j,l}_{y \cdot dep} \geq \hat{t}^{j,(l-1)}_{y \cdot arr} +  \frac{\beta^f - \beta_y^{j, l}}{\min(r_j, \hat{r}_y)} + W_y; \hfill \forall e_j \in \mathcal{E}, \forall s_l \in \mathcal{S}, \forall y \in \mathcal{C}$ \\ \hline
            % C22 & $\mathcal{Z}^{j, l}, \rchi_{i, i'}^{j, l}  \in \{0, 1\}$\\ \hline
        \end{tabular}
\vspace{-0.05in}
\end{table}

\noindent \textbf{Time Constraints:} C15 states that the departure time of an EV from the depot for its sub-trip is a positive quantity.
% \begin{small}
% \vspace{-0.1in}
% \begin{flalign*}
% % \label{eqn:depot_time_setting}
%     \hat{t}^{j, 1}_{d_0\cdot dep} \geq 0, \,\,\, \forall e_j \in \mathcal{E}
% \end{flalign*}
% \vspace{-0.1in}
% \end{small}
An EV can reach a delivery point from a prior delivery or CP, with arrival time constrained by $t_{y \cdot arr} \geq  \sum_{x \in \mathcal{D}} \rchi_{x, y}^{j,l}(t_{x \cdot dep} +  \gamma(x, y)) + \sum_{x \in \mathcal{C}} \rchi_{x, y}^{j, l}(\hat{t}^{j, l}_{x \cdot dep} +  \gamma(x, y))$. This includes the nonlinear terms $\rchi_{x, y}^{j,l} \cdot t_{x \cdot dep}$ and $\rchi_{x, y}^{j, l}\hat{t}^{j, l}_{x \cdot dep}$. 
To linearize them, we introduce decision variables $\Omega_{x,y}^{j,l} = \rchi_{x, y}^{j,l} \cdot t_{x \cdot dep}$ and $\hat{\Omega}_{x,y}^{j,l} = \rchi_{x, y}^{j,l} \cdot \hat{t}_{x \cdot dep}^{j,l}$. 
These assignment operations can be enforced in a MILP setup by incorporating additional constraints, as shown in C16 and C17. The revised constraint for arrival time at a delivery point is provided in C18.
In our setup, an EV arrives at a CP only from a delivery point, and it is captured using C19, which is similar to C18.
The departure time from a delivery point $y$ is determined by Constraint C20, which ensures adherence to the delivery time window and accounts for the package unloading time.
The departure time from a CP $y$ is determined by C21, where $r_j$ and $\hat{r}_y$ represent the charge acceptance rates of the vehicle and charging rate of CP, respectively.
The effective charging rate of vehicle $e_j$ at a CP $y$ can be computed as $\min(r_j, \hat{r}_y)$.
C21 accounts for the arrival time, the time required to recharge the vehicle fully, and any queue waiting time. 
\section{The MILP Formulation}\label{sec:MILP_Formulation}
% \vspace{-0.07in}
The EDRP, as formulated in Eq. (\ref{eqn:obj}), seeks to optimize delivery operations by maximizing the number of successful deliveries while minimizing en-route charging costs. This objective function incorporates normalization factors, $\alpha_1$[unitless] and $\alpha_2$[$\frac{1}{\$}$], as the range of growth of the first and second parts of the objective function are different. The optimization process is subject to constraints detailed in Tables III-VI, ensuring that solutions adhere to practical considerations such as vehicle capacities, time windows, and charging station availability.
\begin{align}
% \vspace{-0.15in}
% \label{eqn:MILP_Formulation}
   & \max \left(
   \alpha_1 \cdot \left(\sum_{e_j \in \mathcal{E}} \sum_{s_l \in \mathcal{S}} \sum_{x \in \mathcal{C} \cup \mathcal{D} \cup \{d_0\}} \sum_{y \in \mathcal{D}, y\neq x} \rchi_{x, y}^{j, l}\right) - \right. \notag \\
   & \left.\alpha_2 \cdot \left(\sum_{e_j \in \mathcal{E}} \sum_{s_l \in \mathcal{S}} \sum_{x \in \mathcal{D}} \sum_{y \in \mathcal{C}\cup \{d_0\}, y\neq x} \rchi_{x, y}^{j, l} \cdot \beta^{j,l}\cdot \theta_y\right)\right) \label{eqn:obj}
   % \textit{s.t.} &  \qquad \quad \text{Constraints expressed in Tables III-VI} 
% \vspace{-0.15in}
\end{align}
However, Eq.(\ref{eqn:obj}) is nonlinear due to the term $\rchi_{x, y}^{j, l} \cdot \beta^{j, l}\cdot \theta_y$. Though $\theta_y$ is a fixed quantity, the term $\rchi_{x, y}^{j, l} \cdot \beta^{j,l}$ is still a binary-continuous product. Therefore, we linearize it by introducing the variable $\lambda_{x, y}^{j, l} = \beta^{j,l} \cdot \rchi_{x, y}^{j, l}$ and the linear objective function is captured in Eq. (\ref{eqn:obj_MILP}) subject to the linearization constraints in Eq. (\ref{Eq:linearity2}) as well as the constraints expressed via Tables III-VI.
% To solve it using MILP this equality can be achieved by adding the below constraints: 
% \begin{small}
% \vspace{-0.04in}
% \begin{align}
% % \label{Eq:linearity}
%     &\lambda_{x, y}^{j, l} \geq 0 \quad\quad \lambda_{x, y}^{j, l} \leq \beta^{j,l} \quad\quad\ \lambda_{x, y}^{j, l} \leq \beta^f \cdot \rchi_{x, y}^{j, l} \notag\\
%     &\lambda_{x, y}^{j, l} \geq \beta^{j,l} - (1 - \rchi_{x, y}^{j, l}) \cdot \beta^f \quad\quad 
%      \quad\quad
%     \forall s_l \in \mathcal{S},\, \forall e_j \in \mathcal{E} \label{Eq:linearity2}
% % \label{Eq:linearity4}
% \end{align}
% \end{small}
\begin{align}
% \vspace{-0.1in}
% \label{eqn:MILP_Formulation}
   & \max \left(
   \alpha_1 \cdot \left(\sum_{e_j \in \mathcal{E}} \sum_{s_l \in \mathcal{S}} \sum_{x \in \mathcal{C} \cup \mathcal{D} \cup \{d_0\}} \sum_{y \in \mathcal{D}, y\neq x} \rchi_{x, y}^{j, l}\right) - \right. \notag \\
   & \left.\alpha_2 \cdot \left(\sum_{e_j \in \mathcal{E}} \sum_{s_l \in \mathcal{S}} \sum_{x \in \mathcal{D}} \sum_{y \in \mathcal{C}\cup \{d_0\}, y\neq x} \lambda_{x, y}^{j, l} \cdot  \theta_y\right)\right)
   \label{eqn:obj_MILP} \\
   &\lambda_{x, y}^{j, l} \geq 0, \quad \lambda_{x, y}^{j, l} \leq \beta^{j,l} \quad\quad\ \lambda_{x, y}^{j, l} \leq \beta^f \cdot \rchi_{x, y}^{j, l} \notag\\
&\lambda_{x, y}^{j, l} \geq \beta^{j,l} - (1 - \rchi_{x, y}^{j, l}) \cdot \beta^f, \quad\quad \forall s_l \in \mathcal{S},\, \forall e_j \in \mathcal{E} 
% \label{Eq:linearity4}
% & \text{Constraints expressed in Tables III-VI}
\label{Eq:linearity2} 
% \vspace{-0.1in}
\end{align}
By setting the battery capacity to $\infty$, the problem addressed in this work can be polynomial-time reduced to the Vehicle Routing Problem (VRP), which is known to be $\mathcal{NP}$-Hard \cite{kallehauge2008formulations}, and the formal proof for the same is provided subsequently.
\begin{theorem}
The problem in Eq. (\ref{eqn:obj}) is $\mathcal{NP}$-Hard.
\end{theorem}
\begin{proof}
CARGO minimizes the cost of delivering items, while maximizing the number of deliveries within specified time windows using EVs while adhering to energy constraints. It also incorporates charge scheduling to minimize charging costs. The problem can be reduced to VRP~\cite{lenstra1981complexity, 9797512}, which optimizes routes for a set of vehicles and customers.
By setting the battery capacity $\beta^f = \infty$, we can relax the energy constraints, allowing each ET to complete its deliveries without recharging. In this case, the problem reduces in polynomial time to a VRP with time windows, which is $\mathcal{NP}$-Hard~\cite{kallehauge2008formulations}. Additionally, by setting $\tau_i^{start} = 0$ and $\tau_i^{end} = \infty$, the problem can be reduced to a simple VRP in polynomial time. Thus, our original problem is at least as hard as VRP. Since VRP is $\mathcal{NP}$-Hard~\cite{lenstra1981complexity}, our problem can be safely deduced as $\mathcal{NP}$-Hard.
\end{proof}

% Due to space constraints, we do not include the proof of intractability.
% Therefore, the MILP for EDRP can be represented using the objective function presented in Eq.(\ref{eqn:obj_MILP}) subject to the constraints listed in Table~\ref{tab:constraint_table} along with Eq. (\ref{Eq:linearity}) to (\ref{Eq:linearity4}).
% \vspace{-0.1in}
\section{Proposed Heuristic Solution}\label{sec:solution_approach}
% \vspace{-0.05in}
Given that EDRP is an $\mathcal{NP}$-hard problem, solving it using the MILP method is computationally intensive. This section introduces a polynomial-time heuristic, called CSA,along with its asymptotic complexity.
% \vspace{-0.1in}
\subsection{Cluster-Sort-Assign (CSA) Heuristic}
CSA finds a cost-optimal delivery route $\mathcal{A}_j$ and a schedule $\mathcal{T}_j$ for each EV $e_j \in \mathcal{E}$ (Algorithm~\ref{algo:CSA}). The route includes the location of delivery and charging, while the schedule specifies the departure time from each location. CSA complies with battery constraints and delivery time windows,
clusters deliveries according to their spatial locations, and the end time of the delivery window. Then, each delivery is assigned to an EV to achieve a local cost-optimal solution.
\vspace{0.05in}

\noindent \textbf{Step 0 (Pre-Processing):} 
While delivering, an EV with a low battery must locate and reach an affordable CP before depletion. 
A brute-force search scales inefficiently with the number of CPs, and selecting a low-cost option further increases complexity.
To optimize this, \textit{Step 0} constructs a K-Dimensional (K-D) tree~\cite{friedman1977algorithm} using latitude, longitude, and unit charging cost of all CPs. Since these dimensions vary in scale, Min-Max normalization is applied. 
The nearest low-cost charging option for any new data point (latitude, longitude, and unit charging cost) is found in logarithmic time~\cite{friedman1977algorithm} by a fast nearest neighbor search in the K-D tree. 
However, deliveries as data points lack a dimension: unit charging cost. To resolve this, the lowest unit charging rate among all CPs is used while querying the K-D tree. The best charging option for each delivery is precomputed and stored at this step.

% Querying the K-D tree, the algorithm pre-computes the nearest low-cost charging point for each delivery with logarithmic time complexity~\cite{friedman1977algorithm}. 

% To complete a delivery, an EV must reach a nearby charging point before its battery is fully depleted. Therefore, when assigning a delivery task, the algorithm must ensure that the residual battery suffices to complete the delivery and reach a nearby charging point.
\setlength{\textfloatsep}{0.1cm}
\setlength{\floatsep}{0.1cm}
\begin{algorithm}[t]
\setstretch{0.85}
    \caption{\textbf{Cluster-Sort-Assign}}
    \label{algo:CSA}
    % \footnotesize
    \small
    \DontPrintSemicolon
    % \scriptsize
    \KwIn{$\mathcal{D}, \mathcal{E}, \mathcal{C}$}
    \KwOut{Delivery trips $\mathcal{A}_j$ and schedule $\mathcal{T}_j$ for each EV $e_j$}

    \tcc{Step 0: PreProcessing}
    Construct a K-D tree using all spatial locations and unit charging costs of CPs.\;
    Identify and store each delivery's nearest cost-optimal CP by searching for its nearest neighbor in the K-D tree.\;

    \tcc{Step 1: Spatio-temporal Grouping}
    $\mathbb{G} \gets ST$-$DBSCAN(\mathcal{D})$\;

    % Sort deliveries $d_i \in g$ in ascending order of their delivery window end time $\tau_i^{end}$; $\forall g \in \mathbb{G}$.\;

    Sort \textit{deliveries} $d_i$ in each group $g \in \mathbb{G}$ in ascending order of their delivery window end time $\tau_i^{end}$.\;
    
    Sort delivery \textit{groups} $g \in \mathbb{G}$ in ascending order of their earliest delivery window end time.\;
    
    \tcc{Step 2: Delivery to EV assignment}
    \For{each $g \in \mathbb{G}$}{
        \For{$d_i \in g$}{
            Initialize the assigned EV for the delivery to $\emptyset$\;
            Initialize min energy cost $\Psi_{min}$ for this delivery to $\infty$\;
            \For{$e_j \in \mathcal{E}$}{
                $x_{last} \gets$ last location of $e_j$\;
                $\Psi \gets $ energy requirement to reach  $d_i$ from $x_{last}$\;
                \If{ $\Psi < \Psi_{min}$}{
                    \If{$e_j$ cannot deliver $d_i$ using its residual battery}{
                        Find the most suitable CP $y \in \mathcal{C}$ from the last location $x_{last}$ using K-D tree.\;
                        Add $y$ to $\mathcal{A}_j$ and send $e_j$ for charging.\;
                        Update the schedule $\mathcal{T}_j$ as per the traveling, charging, and waiting time.\;
                        continue to next EV\;
                    }
                    \If{$e_j$ can reach $d_i$ before $\tau_i^{end}$}{
                        Update the assigned EV for the delivery to $e_j$.\;
                    }
                }
            }
            \If{The assigned EV for the delivery is not $\emptyset$}{
                Let $e_{j'}$ be the assigned EV\;
                Add location of $d_i$ to $\mathcal{A}_{j'}$ and assign $e_{j'}$ to deliver $d_i$\;
                Update the schedule $\mathcal{T}_{j'}$ depending on the traveling time, unloading time and delivery time window.\;
                Update the residual battery of $e_{j'}$\;

            }  
        }
    }
    For all EVs return to the depot.\;
\end{algorithm}

\vspace{0.05in}
\noindent\textbf{Step 1 (Cluster and Sort):} 
% Geographically close delivery points may have different delivery time windows, while deliveries with similar time windows may be far apart. 
Nearby deliveries may have different time windows, while similar time windows may belong to deliveries with distant locations.
As a result, an EV prioritizing deliveries with close time windows may travel longer distances, increasing energy consumption, whereas an EV serving only nearby deliveries risks missing deadlines. 
To balance this space-time trade-off, CSA employs a spatio-temporal clustering technique, ST-DBSCAN~\cite{stdbscan}, to group deliveries based on spatial proximity and the end time of the delivery window $\tau_i^{end}$.
However, each delivery group $g \in \mathbb{G}$ may contain deliveries with varying time windows. Therefore CSA first sorts deliveries within each group by their $\tau_i^{end}$ to ensure timely deliveries. It then orders the groups based on the earliest delivery deadline given by $\min({\tau_i^{end}: d_i \in g})$.
% To prioritize deliveries with earlier deadlines, CSA first sorts the deliveries within each group by their end time. Then sort the groups $g \in \mathbb{G}$ based on the earliest delivery end time, represented as $\min({\tau_i^{end} : d_i \in g})$.

% The clusters are sorted by their earliest delivery end time among all deliveries, then by sorting each cluster's deliveries by $\tau_i^{end}$ [Lines 2-4].

\vspace{0.05in}
\noindent\textbf{Step 2 (Assign):}
    For each delivery, this step selects the most cost-efficient EV capable of completing it. 
    CSA evaluates the energy cost for all EVs to perform a delivery $d_i$ from the sorted delivery groups from Step 1. Beyond cost efficiency, the algorithm ensures the EV can reach the delivery location before the end time and has enough residual battery to reach the nearest CP afterward. 
    The most cost-effective EV meeting time and energy constraints is assigned for delivering $d_i$, and its residual energy, departure time, and trip details are updated. 
    % The most cost-effective EV that meets time and energy constraints is selected for the delivery. The residual energy, departure time, and trip details of the chosen EV are then updated.
    If an EV lacks sufficient battery, it is directed to the nearest CP from its last delivery. Any query for nearby charging is resolved in $O(1)$ time using data precomputed in Step 0.
    % For each delivery, this step searches for the most cost-efficient EV capable of completing it. CSA evaluates the energy cost for all EVs using the sorted delivery groups from Step 1. In addition to cost efficiency, the algorithm checks if the EV can arrive at the delivery location before the end time and has sufficient residual battery to reach the nearest charging point after completing the delivery. If an EV's residual battery is insufficient, it is directed to the nearest charging point from its last assigned delivery location. Among the EVs that meet both time and energy constraints, the most cost-effective EV is chosen for delivery. 
    % The residual energy, departure time, and trip of the chosen EV is updated accordingly. In CSA, any query for nearby charging options is resolved in $O(1)$ time using precomputed data from the K-D tree.
% \end{enumerate}
% \vspace{0.1in}
%%%% Edited until above -- to do complexity analysis below %%%
\vspace{-0.05in}
\subsection{Time Complexity} 
\vspace{-0.05in}
%The generation of K-D trees for the charging stations takes $O(|\mathcal{C}|$ $\log$$|\mathcal{C}|)$ where $|\mathcal{C}|$ is the number of charging stations. Precomputing the K-D tree at \textit{Pre-process} stage (Step 0) enables reusing the results during the \textit{Assign} stage (Step 2). For $|\mathcal{D}|$ delivery points, the above process takes $O(|\mathcal{D}|(\log |\mathcal{C}|))$ time on average, and $O(|\mathcal{D}||\mathcal{C}|)$ time in the worst case. ST-DBSCAN dynamically generates $|\mathbb{G}|$ number of clusters with approximately $\frac{|\mathcal{D}|}{|\mathbb{G}|}$ delivery points per cluster in $O(|\mathcal{D}| \log |\mathcal{D}|)$ time. Once clusters are formed, they are sorted in $O(|\mathbb{G}| \log |\mathbb{G}|)$ time based on their earliest delivery end time. 
% computed in a constant $O(1)$ time. 
The K-D tree for the CPs is created in $O(|\mathcal{C}|$ $\log$$|\mathcal{C}|)$ time, where $|\mathcal{C}|$ is the number of CPs. 
Nearby cheapest charging options for 
$|\mathcal{D}|$ deliveries are precomputed in $O(|\mathcal{D}|$$(\log |\mathcal{C}|))$ time on average, and $O(|\mathcal{D}|$$|\mathcal{C}|)$ time in the worst case. 
% Precomputed results from the K-D tree at \textit{Pre-process} stage are reused in the \textit{Assign} stage.  $|\mathcal{D}|$ delivery points take $O(|\mathcal{D}|$$(\log |\mathcal{C}|))$ time on average for the above process, and $O(|\mathcal{D}|$$|\mathcal{C}|)$ time in the worst case. 
ST-DBSCAN generates $|\mathbb{G}|$  clusters with approximately $\frac{|\mathcal{D}|}{|\mathbb{G}|}$ delivery points per cluster in $O(|\mathcal{D}| \log |\mathcal{D}|)$ time. Then, the clusters are sorted in $O(|\mathbb{G}| \log |\mathbb{G}|)$ time based on their earliest delivery end time. 
\begin{figure*}[!htbp]
% \vspace{-0.15in}
    \centering
    \begin{minipage}[t]{0.32\textwidth}  % Adjusted width
    \centering
    \includegraphics[width=\textwidth]{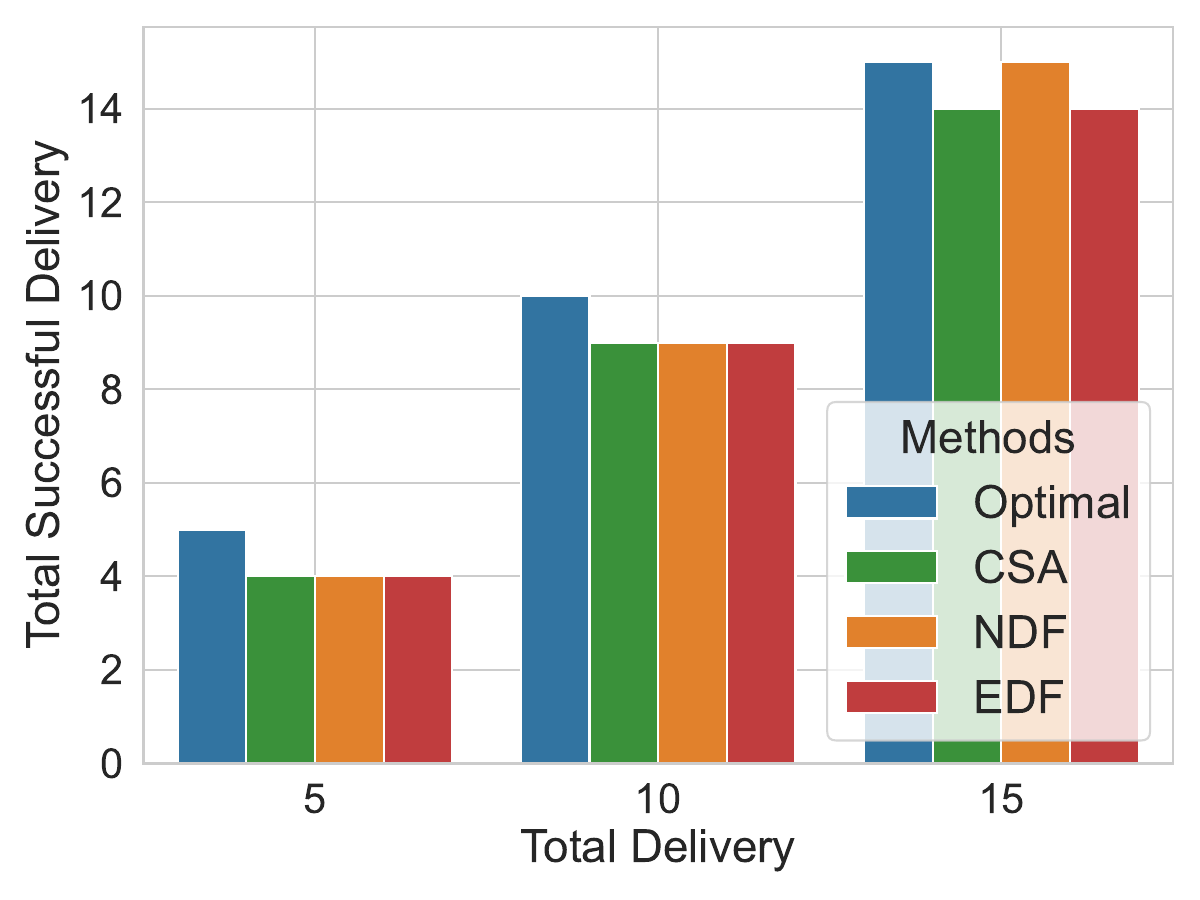}
    \subcaption{No. of successful deliveries \textit{vs.} Total  deliveries.} 
    \label{fig:fig1_opt}
    \end{minipage}
    \hfill
    \begin{minipage}[t]{0.32\textwidth}  % Adjusted width
    \centering
    \includegraphics[width=\textwidth]{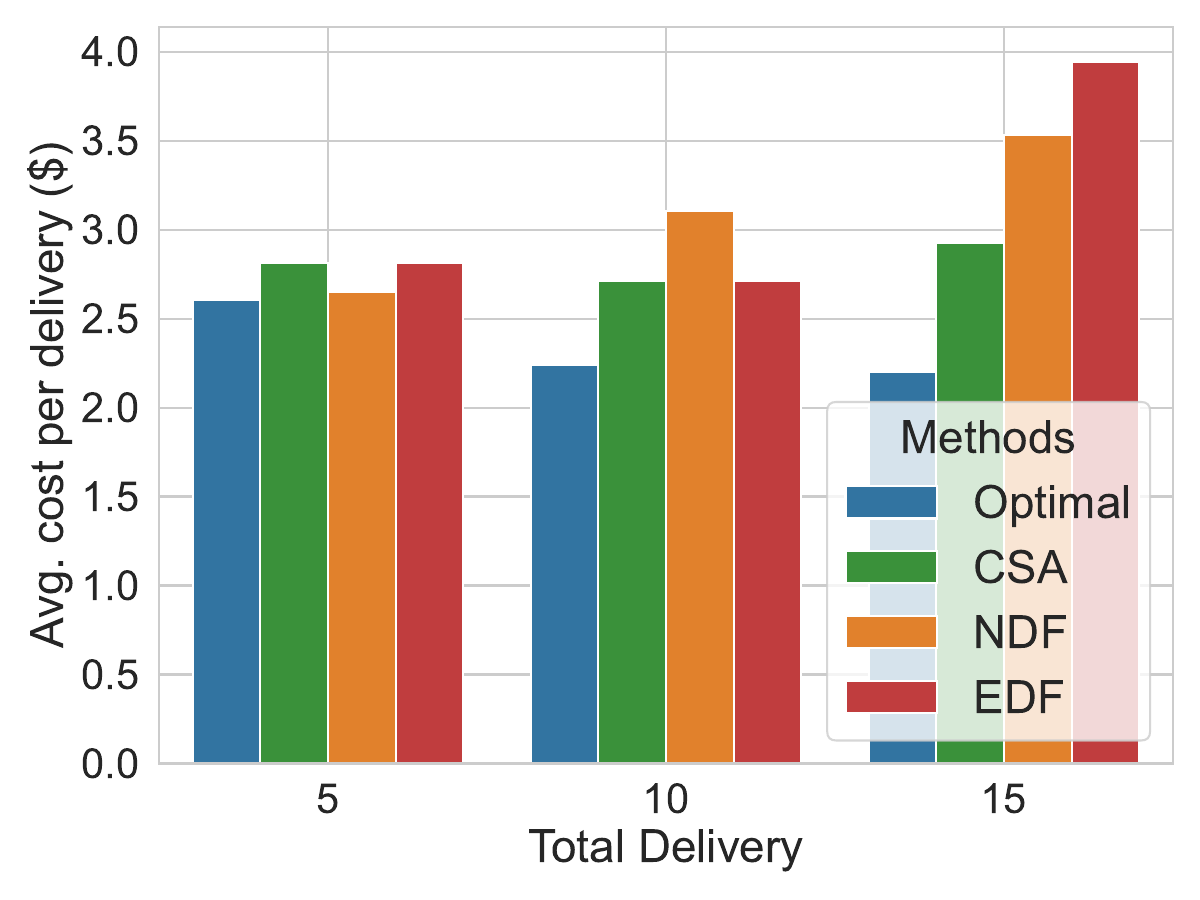}
    \subcaption{Average cost \textit{ vs}. No. of deliveries.} 
    \label{fig:fig2_opt}
    \end{minipage}
    \hfill
    \begin{minipage}[t]{0.32\textwidth}  % Adjusted width
    \centering
    \includegraphics[width=\textwidth]{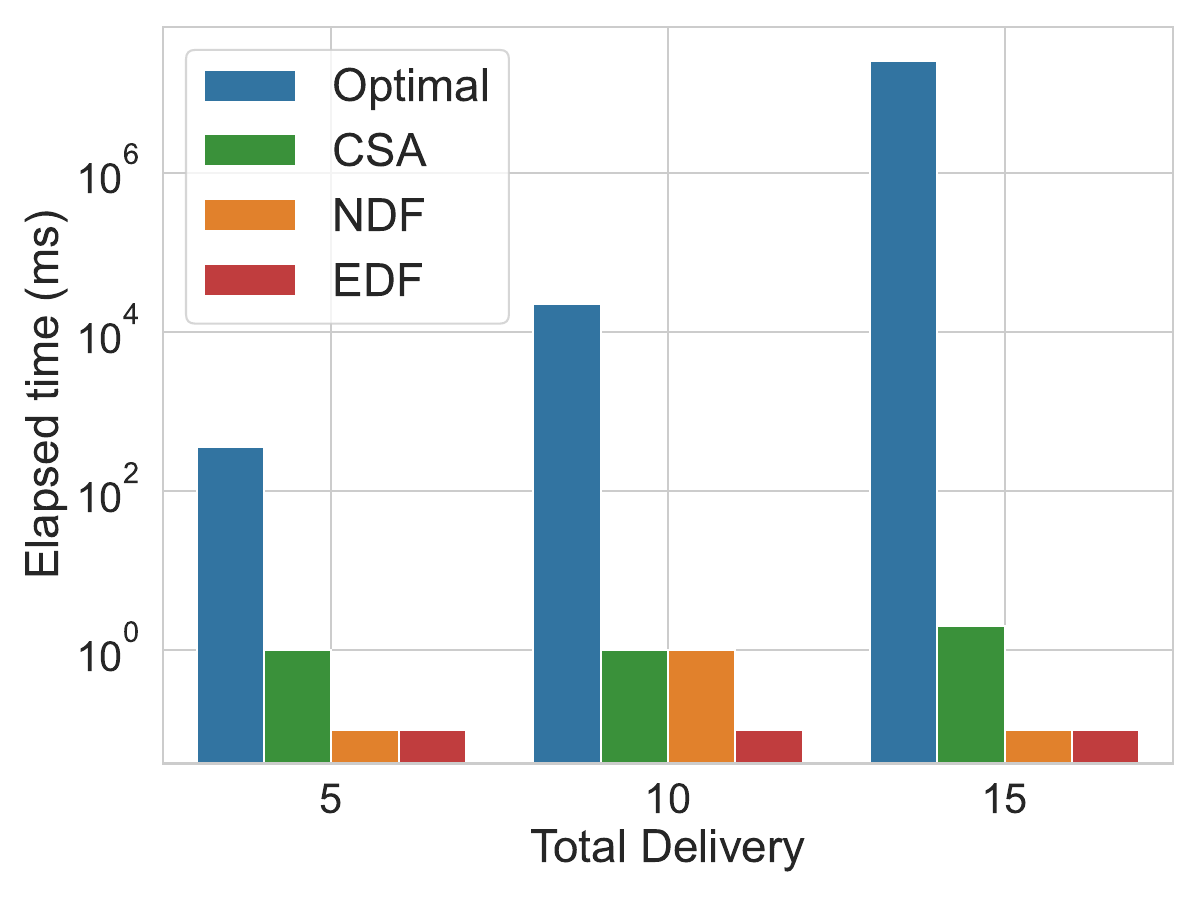}
    \subcaption{Elapsed Time \textit{vs}. Number of deliveries.} 
    \label{fig:fig3_opt}
    \end{minipage}
    % \begin{minipage}[t]{0.24\textwidth}  % Adjusted width
    % \centering
    % \includegraphics[width=\textwidth]{Figures/opt/Total Cost_delivery2ev_ratio_opt.pdf}
    % \subcaption{Total cost  \textit{vs.} No. of deliveries.} 
    % \label{fig:fig4_opt}
    % \end{minipage}
    \caption{Comparison of successful deliveries, average cost, and elapsed time.}
    \label{fig:comparison_with_OPT}
    \vspace{-0.1in}
\end{figure*}
% computed in a constant $O(1)$ time. 
The total time to sort all delivery points in each cluster is $O(\frac{|\mathcal{D}|}{|\mathbb{G}|} \log \frac{|\mathcal{D}|}{|\mathbb{G}|})$. The \textit{Assign} stage takes $O(|\mathcal{D}|$ $|\mathcal{E}|)$ time to find the least energy-expending EVs for deliveries. The closest and cheapest CP is retrieved in $O(1)$ time from the pre-computed results for a charge-decayed EV.  
Therefore, the worst-case time complexity of Algorithm~\ref{algo:CSA} can be expressed as $O(|\mathcal{C}|\log |\mathcal{C}| + |\mathcal{D}||\mathcal{C}| + \dfrac{|\mathcal{D}|}{|\mathbb{G}|}\log {|\mathcal{D}|}{|\mathbb{G}|} + |\mathbb{G}|\log |\mathbb{G}| + |\mathcal{D}|\log |\mathcal{D}|+ |\mathcal{D}||\mathcal{E}|)$, which reduces to  $O(|\mathcal{D}||\mathcal{C}| + |\mathcal{D}|\log |\mathcal{D}|)$, with the assumption that $|\mathcal{C}| <|\mathcal{D}|$, $|\mathcal{E}|<|\mathcal{D}|$, and $|\mathbb{G}|<|\mathcal{D}|$.
\vspace{-0.05in}
\section{Experimental Validation}\label{sec:result}
\vspace{-0.1in}
This section details the experimental setup, covering the environment configuration, dataset characteristics, implementation specifics, and key findings from the comparative analysis between the proposed methods and baseline strategies.

\vspace{-0.12in}
\subsection{Environmental Setup}
\vspace{-0.05in}
\noindent \textbf{Dataset Used:} To benchmark the performance of the proposed scheme, we use the real-world \texttt{JD Logistics} dataset \cite{zheng2024hybrid}. It is collected from operations in Beijing, including delivery requests for purchased goods and customer pickup requests. It includes delivery and pickup requests 
paired with \textit{100} charging stations. 
Each node (customer or charging station) in the dataset provides its geographic coordinates (longitude and latitude). Travel distances and times between nodes may violate the triangle inequality, reflecting real-world routing conditions. EV parameters, such as battery capacity and energy consumption rate per unit distance, are derived directly from the dataset. Instance files include node-level attributes (type, location, demand, time windows, service times) and edge-level data (travel distance and time), offering a realistic and scalable benchmark for evaluating EV routing and charging strategies.
The dataset models all EVs with identical charge consumption and all CPs with uniform recharging speed and cost. In contrast, our model introduces heterogeneity: EVs have random consumption rates, and CPs have recharging rates drawn from the 50–350~\textit{kW} range of Level~3 DC chargers~\cite{evbox2023level3}, with unit costs randomly selected between \$0.4 and \$0.6 per \textit{kWh}.

\vspace{1pt}
\noindent \textbf{Implementation Environment:} The MILP-based optimal solution is implemented using \texttt{Python 3.8.3} with the \texttt{Gurobi 12.0 solver}, chosen for its effectiveness in handling large optimization problems \cite{gurobipy}. 
% \texttt{Gurobi} experiments were conducted on a CPU cluster equipped with dual $64$-core \texttt{AMD EPYC Milan 7713} processors and \textit{64} \textit{GB} of \texttt{RAM}, utilizing $128$ logical cores. 
The implementation is publicly available at our repository \cite{CARGO2025}. CSA is also implemented in \texttt{Python}, modeling EVs and charging stations as separate classes. CSA is benchmarked against two baselines: (1) Earliest Deadline First (EDF), which prioritizes deliveries based on their start times \cite{faggioli2009implementation}, and (2) Nearest Delivery First (NDF) \cite{sarjono2014determination}, which assigns deliveries based on proximity to the depot. Both baselines ensure feasibility by respecting time and battery constraints. All algorithms are executed on an \texttt{Intel Core i7-9750H} CPU with $16$ GB of RAM. The \texttt{Gurobi}-based implementation uses $12$ CPU threads for parallel execution.

\subsection{Results and Analysis}
% \vspace{-0.05in}
\noindent \textbf{Comparison with Optimal:} 
In our first experiment, we evaluate the performance of the MILP-based solution (referred to as Optimal) in comparison to CSA and the baselines EDF and NDF. We consider a delivery scenario ranging from $5$ to $15$ deliveries, with a delivery-to-EV ratio of $\frac{|\mathcal{D}|}{|\mathcal{E}|} = 5$, and $5$ CPs.
Fig.~\ref{fig:fig1_opt} illustrates the number of successful deliveries. Optimal completes all deliveries within their respective time windows, while the other methods generally fall short by one delivery. Note that the execution time of the Optimal solution increases exponentially with the number of deliveries; the Gurobi-based implementation becomes practically infeasible to execute, hence we restrict our analysis to 15 deliveries.
% \textcolor{red}{AS: Maybe we can add one line stating that for larger test cases it wasn't executing and taking more time?}

We assess the efficiency of assigning EVs to low-cost charging options on the delivery route based on the average cost incurred per delivery to complete successful deliveries. 
% \textcolor{red}{AS: Just a thought, is average a good indicator here?}
\begin{figure*}[t]
    \centering
    \begin{minipage}[t]{0.32\textwidth}  % Adjusted width
    \centering
    \includegraphics[width=\textwidth]{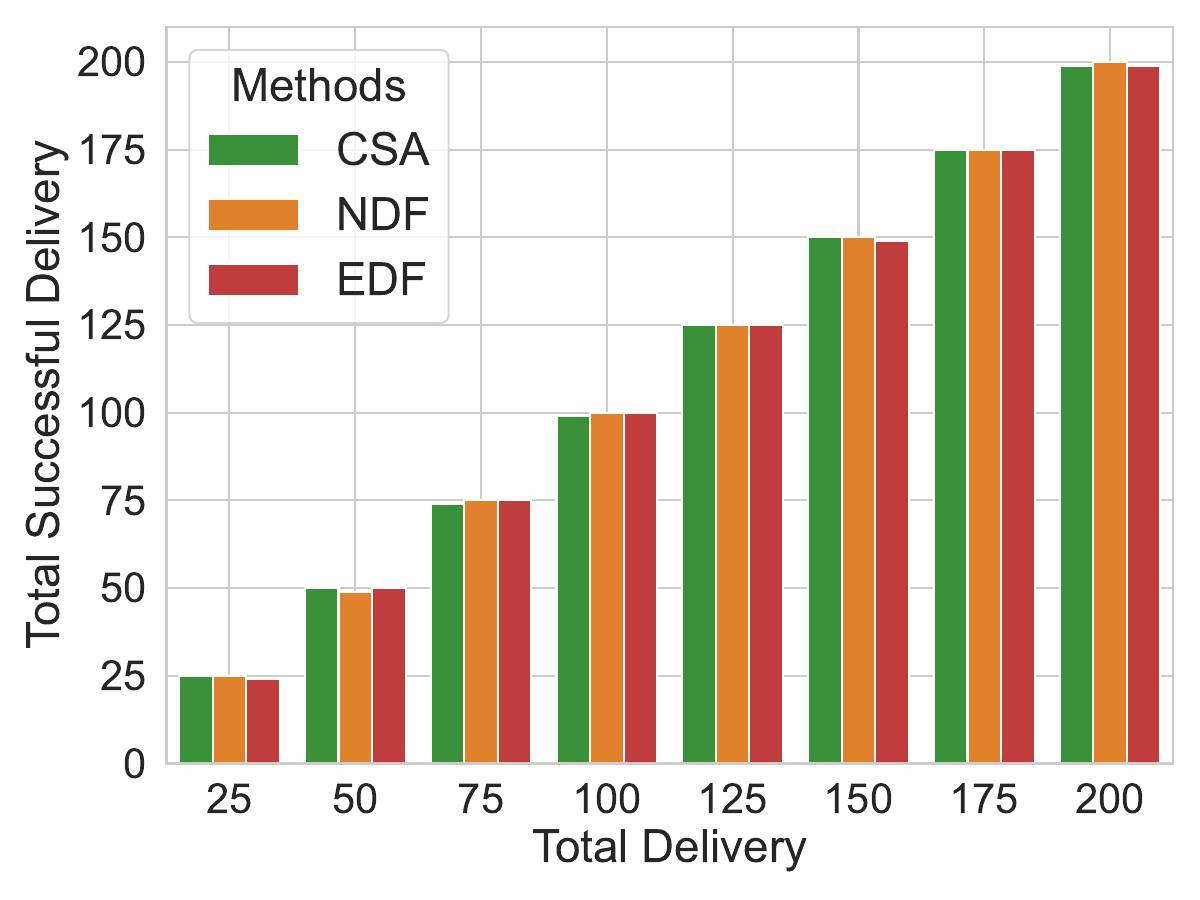}
    \subcaption{Number of successful deliveries \textit{vs.} Total  Deliveries.} 
    \label{fig:scale_1}
    \end{minipage}
    \hfill
    \begin{minipage}[t]{0.32\textwidth}  % Adjusted width
    \centering
    \includegraphics[width=\textwidth]{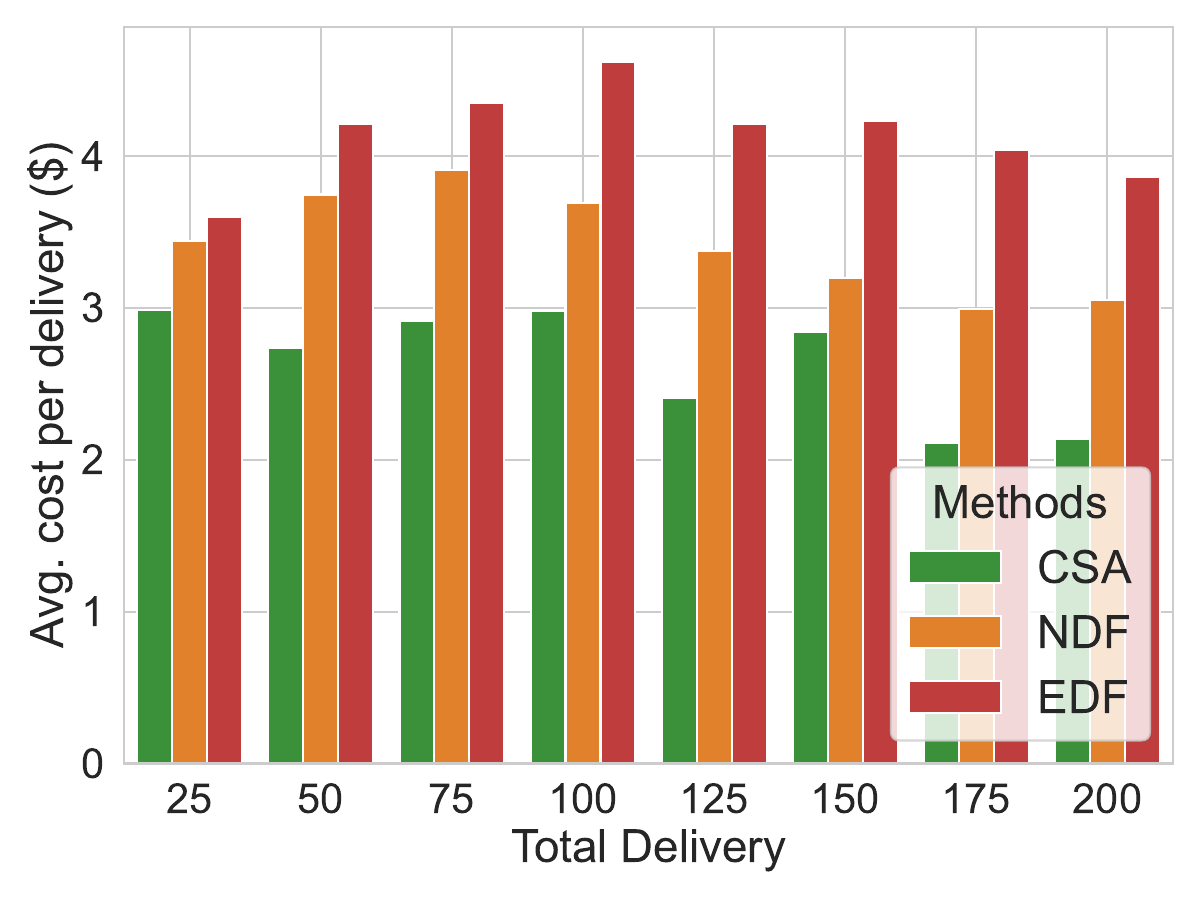}
    \subcaption{Average cost  \textit{vs.} No. of deliveries.} 
    \label{fig:scale_2}
    \end{minipage}
    \hfill
    \begin{minipage}[t]{0.32\textwidth}  % Adjusted width
    \centering
    \includegraphics[width=\textwidth]{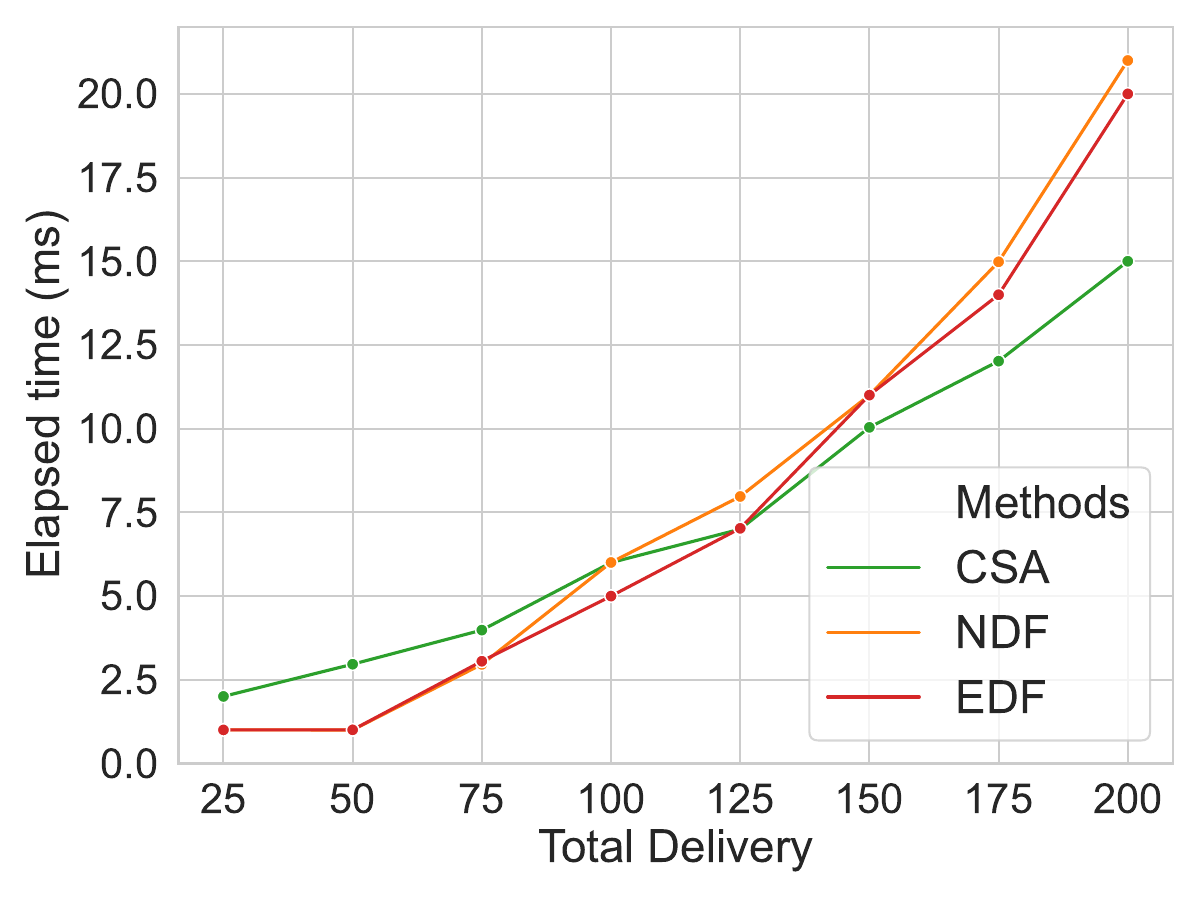}
    \subcaption{Elapsed Time \textit{vs.} No. of deliveries.} 
    \label{fig:scale_3}
    \end{minipage}
    \caption{Comparison of successful deliveries, average cost, elapsed time, by varying delivery to EVs ratio.}
\label{fig:scalability_test}
    \vspace{-0.1in}
\end{figure*}
Fig.~\ref{fig:fig2_opt} presents the average cost per successful delivery, showing that the Optimal consistently incurs the lowest cost, which is expected behavior. 
For the smallest test case (5 deliveries), we observe that CSA cannot effectively leverage clustering and travels longer distances, leading to a higher average cost, while NDF achieves a low travel cost by selecting the nearest deliveries.
Since EDF focuses on meeting early deadlines rather than minimizing route distance, it typically consumes more energy than NDF. However, as the cost depends on the product of energy consumed and the unit charging cost, EDF may sometimes incur a lower total cost, particularly in smaller instances, if it charges at CPs with lower unit prices. We observed such a case in the 10-delivery instance.
% For the smallest test case (5 deliveries), CSA performs similarly to EDF and slightly worse than NDF. 
% \textcolor{red}{AS: Some intuition on why this happened? It may be case-specific, but it's worth looking into.}
% However, as the number of deliveries increases, CSA outperforms both EDF and NDF, achieving lower average costs.
% \textcolor{red}{AS: Should we also write one line about NDF and EDFs comparative behavior?}
Finally, Fig.~\ref{fig:fig3_opt} illustrates total execution time. Optimal incurs exponentially higher computation time due to its brute-force nature. 
While CSA is more efficient than Optimal, CSA runs slower than EDF and NDF in small scenarios due to the overhead of K-D tree construction and clustering.
% \textcolor{red}{AS: One line on the reasoning behind this?}

% However, the real-world scenarios are way larger than this setup and we evaluate the scalability of 
% The performance of CSA is evaluated against the MILP-based solution (referred to as Optimal), with comparative results presented in Fig.~\ref{fig:comparison_with_OPT} for delivery scenarios ranging from 5 to 15 deliveries, using 2 EVs and 2 charging points. Fig.~\ref{fig:fig1_opt} illustrates the number of successful deliveries, where both CSA and OPT complete all deliveries within their respective time windows. However, as the sample size increases, CSA's performance may deviate, as shown in Figs.~\ref{fig:fig1} and \ref{fig:fig9}. The experiment is restricted to these parameters due to the exponential computation time required for more significant instances, even on a cluster machine. Fig.~\ref{fig:fig2_opt} highlights the total execution time for EV-to-delivery assignments, where OPT incurs exponentially higher computational costs due to its brute-force approach. At the same time, CSA offers a more efficient alternative. Regarding cost efficiency, both methods yield similar total and average costs for more minor test cases with 5 deliveries. However, as the number of deliveries grows, OPT outperforms CSA, with the cost gap widening, as evidenced by Figs.~\ref{fig:fig3_opt} and \ref{fig:fig4_opt}.

\vspace{1pt}
\noindent \textbf{Scalability Tests:} 
% As observed in the previous experiment, the execution time of the Optimal solution increases exponentially with the number of deliveries, the Gurobi-based implementation becomes practically infeasible when using a small number of parallel threads. Therefore, 
We conduct a scalability study for CSA and compare it with NDF and EDF due to an exponential increase in execution time of the Optimal with larger problem sizes.

In Fig.~\ref{fig:scalability_test}, we report the number of successful deliveries, average cost, and elapsed time as the number of delivery tasks increases from $25$ to $200$ in increments of $25$, assuming both the number of CPs and the delivery-to-EV ratio are fixed at $5$. 
Figs.~\ref{fig:scale_1} and~\ref{fig:scale_2} illustrate the number of successful deliveries and the average delivery cost, respectively, across different methods. CSA consistently achieves a comparable number of successful deliveries while maintaining the lowest average cost among all methods. This performance is attributed to CSA's cost-aware joint decision-making for routing and charging, in contrast to the NDF and EDF strategies, which rely on nearest and earliest delivery heuristics. Experiments with other combinations of CPs and delivery-to-EV ratios show similar patterns and are omitted to avoid redundancy.

Fig.~\ref{fig:scale_3} shows that execution time increases with the number of deliveries across all methods. Although CSA incurs additional overhead from K-D tree construction and clustering, it benefits from faster charging station searches using the constructed tree. This efficiency enables CSA to outperform EDF and NDF in total execution time for larger delivery scenarios. Notably, CSA exhibits slower growth in execution time and achieves the lowest execution time among all methods once the number of deliveries exceeds $125$.

\begin{figure}[t]
    \centering
    \includegraphics[width=0.9\linewidth]{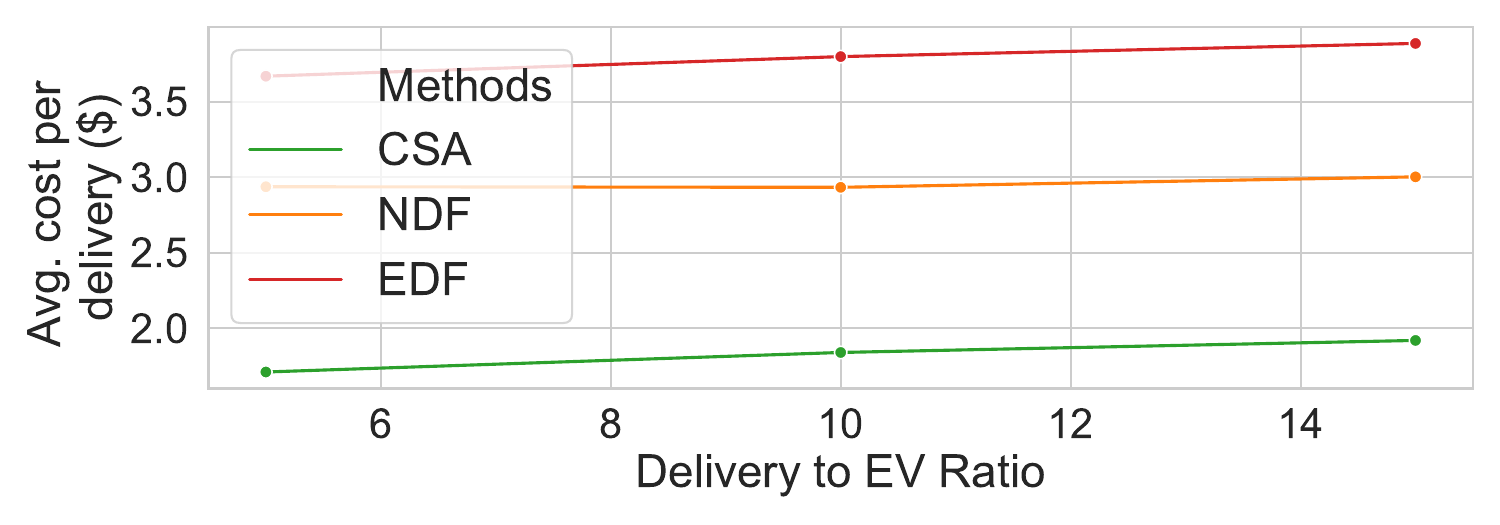}
    \vspace{-0.02in}
    \caption{Avg. cost with varying delivery-to-EV ratio.}
    \label{fig:varying_ratio}
    % \vspace{-0.05in}
\end{figure}

In Fig.~\ref{fig:varying_ratio}, the delivery-to-EV ratio is varied from $5$ to $15$, with the number of deliveries fixed at $200$ with CPs set to $50$. As the ratio increases (i.e., fewer EVs), each EV travels longer distances to meet delivery deadlines, resulting in higher average delivery costs. CSA consistently achieves the lowest average cost by jointly optimizing: (1) delivery routes to minimize energy consumption, and (2) charging assignments by selecting cost-efficient CPs near the routes. In contrast, EDF prioritizes early deadlines without route optimization, resulting in higher energy usage and the most significant recharging cost. NDF improves routing but lacks cost-aware CP assignments, resulting in higher overall costs compared to CSA.

\begin{figure}[t]
    \centering
    \includegraphics[width=0.9\linewidth]{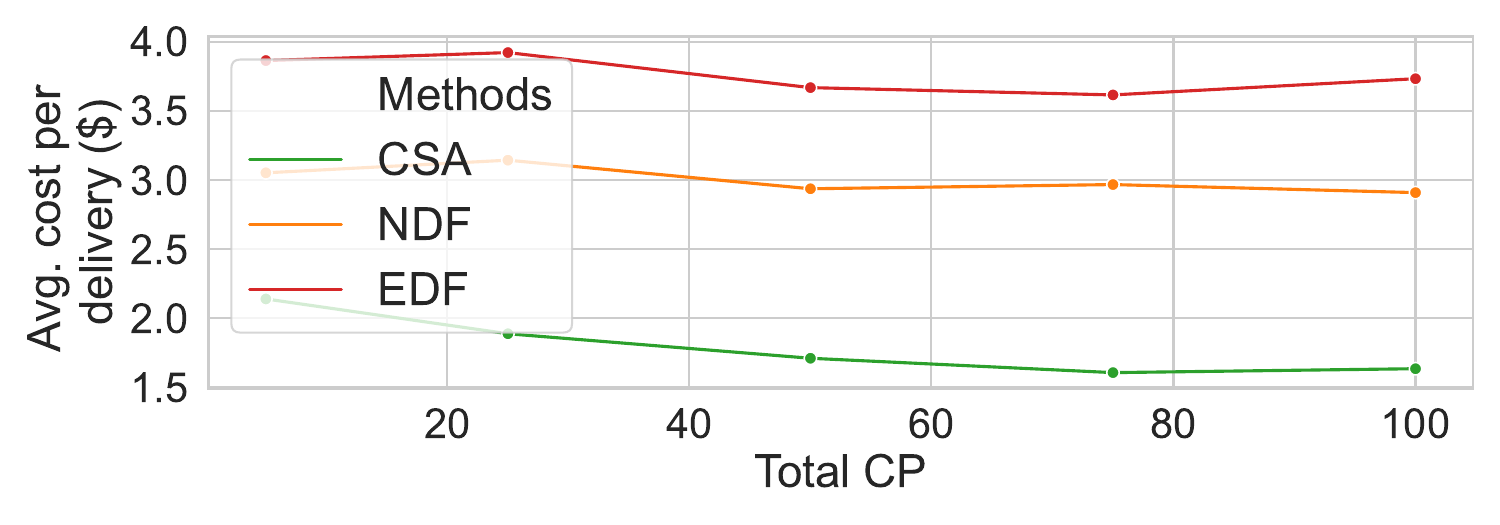}
    \vspace{-0.02in}
    \caption{Avg. cost with varying the number of CP.}
    \label{fig:varying_total_CP}
    % \vspace{-0.05in}
\end{figure}

In Fig.~\ref{fig:varying_total_CP}, we vary the total number of CPs from $5$ to $100$, while keeping the number of deliveries fixed at $200$ and the delivery-to-EV ratio at $5$. Although charging costs are randomly drawn from a fixed range, increasing the number of CPs improves the chances of accessing a nearby low-cost option. CSA capitalizes on this, resulting in a decreasing trend in average charging costs as CP availability increases.


\vspace{-0.05in}
\section{Conclusion}\label{sec:cnls}
% \vspace{-0.05in}
This work presents the \textsc{CARGO} framework as an effective solution to the EV-based Delivery Route Planning (EDRP) problem, addressing key challenges in routing, charging logistics, and time-window compliance. By providing an exact solution and computationally efficient heuristic, \textsc{CARGO} balances delivery efficiency and operational cost. Compared to EDF and NDF, the heuristic achieves charging cost reductions of $\gtrapprox 39\%$ and $\gtrapprox 22\%$, respectively, underscoring its practical value for scalable, cost-effective EV-based delivery systems.

% This work presents a novel solution to the EV-based Delivery Route Planning (EDRP) problem, addressing the critical challenges of route optimization, charging logistics, and time-window adherence for last-mile delivery.  
% The proposed method effectively balances delivery efficiency and operational costs by introducing an optimal MILP solution and a time-efficient heuristic approach.
% On average, the baseline heuristics, EDF and NDF, incur 71.22\% and 62.38\% higher delivery costs than our CSA heuristic. Additionally, CSA achieves 6\% and 8.2\% more successful deliveries than EDF and NDF, respectively.

CARGO demonstrates strong performance in reducing operational costs, adhering to time windows. However, several enhancements remain for future exploration. First, integrating \textit{real-time traffic data} could help mitigate delays caused by congestion and unforeseen events. Second, extending \textsc{CARGO} to support \textit{multi-depot, multi-commodity} scenarios would better reflect the complexity of real-world delivery systems. Third, incorporating \textit{partial charging} strategies in place of full recharges could offer greater flexibility. Finally, optimizing schedules to \textit{reduce idle times} from early EV arrivals would further improve operational efficiency and fleet utilization.

\section*{Acknowledgments}
% The NSF grant acknowledgement details are hidden due to double-blind submission.
This work was supported by the NSF grants: "Cyberinfrastructure for Accelerating Innovation in Network Dynamics" (CANDY) under award \# OAC-2104078, and  ``Satisfaction and Risk-aware Dynamic Resource Orchestration in Public Safety Systems" (SOTERIA) under award \# ECCS-2319995.

\newpage
\bibliographystyle{ieeetr}
\bibliography{sample-base}

\end{document}